\pdfoutput=1

\documentclass[11pt]{article}

\usepackage{algorithm}
\usepackage{algorithmic}

\usepackage[]{acl}
\usepackage{graphicx}
\usepackage{tabularx}
\usepackage{amsfonts}
\usepackage{amssymb}
\usepackage{placeins}
\usepackage[normalem]{ulem}
\usepackage{booktabs} 
\usepackage{float}   
\usepackage{fontawesome5}
\usepackage{times}
\usepackage{latexsym}
\usepackage{amsmath} 
\usepackage{amsthm}

\usepackage{subcaption}

\newcommand{\update}[1]{\textcolor{black}{#1}}

\usepackage[T1]{fontenc}
\newtheorem{proposition}{Proposition}
\newtheorem{assumption}{Assumption}
\newtheorem{definition}{Definition}

\usepackage[utf8]{inputenc}
\usepackage{enumitem}
\setlist[itemize]{noitemsep, nolistsep}

\usepackage{microtype}

\usepackage{inconsolata}

\title{\update{LLMs Enable Bag-of-Texts Representations for Short-Text Clustering}
} 

\author{I-Fan Lin \\
  Leiden University \\
  Leiden,  The Netherlands \\
  \texttt{\fontsize{9pt}{10pt}\selectfont i.lin@liacs.leidenuniv.nl} \\\And
  Faegheh Hasibi \\
  Radboud University \\
  Nijmegen, The Netherlands\\
  \texttt{\fontsize{9pt}{10pt}\selectfont faegheh.hasibi@ru.nl} \\\\\And
  Suzan Verberne \\
  Leiden University \\
  Leiden,  The Netherlands\\
\texttt{\fontsize{9pt}{10pt}\selectfont s.verberne@liacs.leidenuniv.nl} \\}

\begin{document}
\maketitle
\begin{abstract}
In this paper, we propose a training-free method for unsupervised short text clustering that relies less on careful selection of embedders than other methods.
In customer-facing chatbots, companies are dealing with large amounts of user utterances that need to be clustered according to their intent. In these settings, no labeled data is typically available, and the number of clusters is not known.
Recent approaches to short-text clustering in label-free settings incorporate LLM output to refine existing embeddings. 
While LLMs can identify similar texts effectively, the resulting similarities may not be directly represented by distances in the dense vector space, as they depend on the original embedding.
We therefore propose a method for transforming LLM judgments directly into a bag-of-texts representation in which texts are initialized to be equidistant, without assuming any prior distance relationships. 
Our method achieves comparable or superior results to state-of-the-art methods, but without embeddings optimization or assuming prior knowledge of clusters or labels. Experiments on diverse datasets and smaller LLMs show that our method is model agnostic and can be applied to any embedder, with relatively small LLMs, and different clustering methods. We also show how our method scales to large datasets, reducing the computational cost of the LLM use.
The flexibility and scalability of our method make it more aligned with real-world training-free scenarios than existing clustering methods. Our source code is available here: \url{https://anonymous.4open.science/r/BoT_vector-2E0C/README.md}
\end{abstract}

\section{Introduction}

\begin{figure}[t]
        \centering
        \includegraphics[width=\columnwidth]{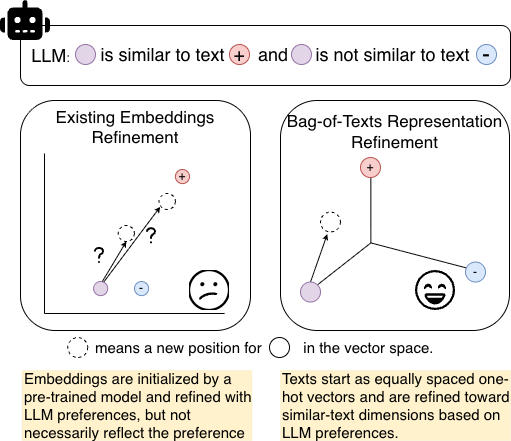}
        \caption{illustration of our proposed method. Left: traditional approach using the LLM output to refine the embeddings space. Right: our approach using the LLM output to directly construct a bag-of-texts space.}
        \label{fig:idea}
\end{figure}

Short text clustering is an NLP task that automatically groups unlabeled data into clusters and is widely used in practical applications for text classification and new category detection. 
For example, in conversational understanding, user intents in real-world scenarios are dynamic and continuously evolving \citep{liang2024survey}, and relying on expert labeling can be costly.
The short-text clustering task enables companies to efficiently organize tens of thousands of client requests in real time. Other downstream tasks, such as mining topics and opinions from online platforms, also rely on effective clustering \citep{wu2024survey}.

The short-text clustering task is often approached as a \textbf{Generalized Category Discovery (GCD) scenario} \citep{vaze2022generalized}, where limited labeled data and substantial unlabeled data are used to cluster all unlabeled samples, which include both known and unknown classes. 
Recently, other studies have explored an \textbf{emerging data scenario}, where no labels are provided at all \citep{de2023idas,viswanathan2024large, zhang2023clusterllm, lin2025spill}. This scenario reduces reliance on human annotation and shows greater flexibility.

Regardless the setting, the classic approach is to train an embedder with contrastive learning \citep{zhang2021supporting}. \update{In recent years, most methods have been designed to use LLMs to guide the optimization \citep{zhang2023clusterllm,zou2025glean}}. Many LLM-guided training-free approaches have been proposed as well to enable easier domain adaptation and reduce the need for fine-tuning \citep{de2023idas,viswanathan2024large,lin2025spill}. Although these training-free approaches lead to substantial improvement, they only partially capture LLM-derived similarities and remain heavily dependent on the original embeddings. The resulting representation may still not reflect LLM preference, despite this being a strong signal. Hence our objective: to construct text representations that directly incorporate LLM similarity estimations into the vector space.

The core idea of our method (Figure~\ref{fig:idea}) is to construct new representations (referred to as bag-of-texts vectors) such that texts belonging to the same cluster get more similar representations. Our method consists of two stages: we first construct bag-of-texts (BoT) representations based on representative texts, and then iteratively refine them by LLM guidance. 
In summary, the contributions of our method are as follows:
\begin{itemize}
    \item \textbf{A Strategy for Encoding LLM Preferences} Unlike prior work that refines existing embeddings, we introduce a Bag-of-Texts (BoT) representation for encoding LLM outputs.

    \item \textbf{Flexible:} Our method can be added to any embedder and does not require fine-tuning, enabling easier adaptation to new domains.
    \item \textbf{Low-resource:} Our method reduces the need for human labeling, can be implemented using relatively small, zero-shot models, and achieves results competitive with labeled or fine-tuned approaches.
\end{itemize}


\section{Related Work}

\paragraph{Short text clustering}
Generalized Category Discovery (GCD) is often studied for this task \citep{liang2024actively, liang2023clusterprompt, zou2025glean}. These works use limited labeled text and substantial unlabeled text to learn clustering representations. However, these studies use train–test splits within a single dataset, which implicitly assumes that new categories are similar to the known ones. In practice, emerging data may show topic drift. Recently, with the advancement of LLMs, many studies have explored the emerging data scenario \citep{zhang2023clusterllm, lin2025spill}, where no labels are available. We follow this unsupervised scenario to support real world applications.

\paragraph{LLM-Guided Contrastive Learning for Short Text Clustering}

With the advancement of LLM, recent work has begun to incorporate LLM feedback into the optimization process. \citet{zhang2023clusterllm} use LLMs to construct hard triplets to derive training pairs. \citet{liang2024actively} identify samples in ambiguous regions and use LLMs to reassign them to new clusters, which are then used as training data. \citet{zou2025glean} proposed a unified framework that actively learns from diverse and quality-enhanced LLM feedback, including instance-level, category description, and uncertain-instance alignment. However, these methods either rely on knowing the number of clusters, proprietary models, or fine-tuning. Our method does not have these requirements.
\paragraph{LLM-Guided Training-Free Approaches for Short Text Clustering}
Training-free approaches use a frozen pre-trained embedder and use LLMs in two ways. The first category is input text enrichment: \citet{de2023idas} select prototypical utterances and then use an LLM to generate a set of descriptive utterance labels for these prototypes. \citet{viswanathan-etal-2024-large} use LLMs for few-shot in-context learning to improve texts by generating key phrases for each text. The second category is embedding post-processing: \citet{lin2025spill} use LLMs for selection rather than text enrichment. They first encode all of the texts, and then use the LLM to select similar pairs for each text. Each text embedding is then obtained by average pooling its similar pairs.  These methods have in common that they refine original embeddings. In contrast, we construct a new vector using LLM feedback.

\section{Bag-of-Texts Theoretical Justification}

\subsection{Problem Definition}
The task is to partition $D^{test}$ into $K$ clusters. $K$ is treated as prior knowledge in most existing work \citep{de2023idas,zhang2023clusterllm, feng2024llmedgerefine, zou2025glean}, and many approaches incorporate $K$ when refining embeddings. Our proposed method does not make any assumption on the knowledge of $K$ thus can be implemented without knowing the number of clusters. For evaluation, however, we incorporate $K$ for fair comparison to previous methods.

In line with prior work, we address two scenarios: (i) the emerging data scenario and (ii) the Generalized Category Discovery (GCD) scenario. In the emerging data scenario \citep{zhang2023clusterllm,lin2025spill}, the only available resource is a collection of unlabeled texts $D^{test} \triangleq \{x_{i}\}_{i=1}^{N}$, where each $x_{i} \in D^{test}$ corresponds to a short text and $N$ is the size of the test dataset. The dataset is used as a whole for clustering and evaluation, without any train-test split. In contrast, the GCD scenario typically has access to additional data, typically including some \textit{labeled} examples \citep{liang2024actively,zou2025glean}. 
As opposed to prior work, to align the GCD scenario with the objective of low-resource settings, we restrict the availability of additional data to only an \textit{unlabeled} dataset $D^u \triangleq \{x_{i}\}_{i=1}^{N_u}$.

For notation convenience, we define the full dataset as $D \triangleq D^{test} \cup D^u$ to simplify the explanation of our proposed method. Note that $D = D^{test}$ in the emerging data scenario.

\subsection{Embedding Refinement via Bag-of-Texts}

Existing LLM-guided short-text clustering approaches often refine embeddings either directly using contrastive learning \cite{zhang2023clusterllm}, indirectly by text enrichment \cite{de2023idas, viswanathan2024large}, or through embedding post-processing \cite{lin2025spill}.
While contrastive optimization enforces inequality through gradient updates, it requires training on new samples, which makes it impractical for low-resource settings.
Text enrichment approaches, on the other hand, can perform for low-resource scenarios but lack explicit guarantees for pulling similar texts closer and pushing dissimilar ones apart. \update{Embedding post-processing approaches heavily depend on the embedder used.} 

Our novel bag-of-texts  representation combines the strengths of these approaches and directly enforces inequality through text-based LLM outputs, without requiring training data or expensive gradient-based updates of embedders. 
Our approach mimics contrastive learning by reconstructing the vector space using a bag-of-texts representation and updating this representation using LLM guidance. 
Using  bag-of-texts representation, all texts start with equivalent distances in the vector space, and LLM output about similarity of texts to each other is directly translated to the vector representation, moving each text toward the correct region of the space; see Figure \ref{fig:idea} for illustration.

\textbf{Formally}, for a text $x$, let $l_{x, x^{+}}$ and $l_{x, x^{-}}$ denote similarities between $x$ and a positive and negative example, and $l^{'}_{x, x^{+}}$ and $l^{'}_{x, x^{-}}$ be the corresponding similarities after embedding refinement.
The positive and negative examples are identified by a function (e.g., an LLM or a human), which determines whether the text $x$ belongs to the same cluster as example $x^{+}$ (positive)  or to a different cluster from the example $x^{-}$ (negative).
Ideally, our refinement approach should satisfy $l^{'}_{x, x^{+}} > l^{'}_{x, x^{-}}$. Below, we formally explain how our bag-of-texts representation achieves this goal.

\begin{assumption}\label{assum_partition_cluster}

We assume that a function (e.g., an LLM or a human) makes a selection based on the following rules.

\begin{itemize}

\item Similarity: For any texts $x_i$ and $x_j$ with $i\neq j$, $x_i$ and $x_j$ are similar if they belong to the same cluster $S_k$, and not similar if they belong to different clusters $S_k$ and $S_h$ with $k \neq h$.

\item \update{Distinct Clusters: Each text is assigned to exactly one cluster $ S \in \mathcal{S}$.} 

\end{itemize}
\end{assumption}

\update{Note that $\mathcal{S}$ in Assumption~\ref{assum_partition_cluster} refers to the clusters defined by the function; there is no knowledge of a ground truth. We assume distinct clusters as is conventional in prior work \citep{an2024generalized,lin2025spill}. 
}

\begin{definition}[Bag-of-Texts]
Let $g(x,D)$ be a function that takes a text $x$ and a set of texts $D$, and outputs a bag-of-text representation $\mathbf{z} \in \mathbb{R}^{|D|}$. In this representation, the entry corresponding to each text in $D$ that belongs to \update{the same cluster} as $x$ is assigned a value of 1, while all other entries are 0.
\end{definition}

\begin{proposition}
Given \textit{Assumption \ref{assum_partition_cluster}}, for any $x_i, x_j \in D$ and their corresponding bag-of-texts representation $\mathbf{z}_i, \mathbf{z}_j$, the function $g(x,D)$ guarantees that $l^{'}_{x, x^{+}} > l^{'}_{x, x^{-}}$. 
\end{proposition}

The proof of the proposition builds on \textit{Definition~1} that if two texts $x_i, x_j \in D$ belong to the same cluster, their bag-of-texts representations $\mathbf{z}_i$ and $\mathbf{z}_j$ are identical. Considering cosine similarity as an example similarity function, $\text{cosine}(\mathbf{z}_i, \mathbf{z}_j) = 1$ if they are identical, and  $\text{cosine}(\mathbf{z}_i, \mathbf{z}_j) = 0$ if they are dissimilar.

\subsection{Adaptation to Real-world Practice}
Although the proposition holds in the idealized setting, in practice the similarity between texts is typically determined using LLMs,
which may introduce errors, and scanning the entire dataset for each $x$ is computationally expensive.
We outline the modifications needed for practical implementation. The detailed analysis is in appendix \ref{rig_analysis}.

\paragraph{Memory constraints.}
Large datasets lead to high-dimensional bag-of-text vectors. To address these issues, we construct bag-of-texts vectors using a representative subset $D^{r} \subseteq D$ with  $|D^{r}| \triangleq d$ , rather than all texts. This dimensionality reduction is possible because it relies on the assumption that texts within the same cluster are similar; therefore, it is not necessary to use all texts in $D$ to represent bag-of-texts vectors. Instead, the bag-of-texts vectors of the remaining texts can be represented via the vectors of the selected representatives. 
\paragraph{Computation constraints.}
Selecting similar texts requires scanning the entire dataset for each text, which is computationally expensive with an LLM. To reduce computation, we use a pre-trained embedder to retrieve a small set of $m$ candidate texts that are likely to be similar, as supported by prior work \citep{lin2025spill,zou2025testnuc}.
\paragraph{Selection uncertainty.}
During automatic selection (e.g., by LLM)
inconsistent preferences can occur and violate \update{Assumption \ref{assum_partition_cluster}}. For example, $(x_i,x_j)$ may be identified as belonging to the same cluster, $(x_i,x_l)$ as different, yet $(x_j,x_l)$ as the same. Also, in practice, each selection is made with some inherent confidence. To reduce these inconsistencies, we perform iterative update and assign the mean of the selected bag-of-texts vectors rather than a fixed value of 1.

\section{Computational Method}
Our method aims to construct Bag-of-Texts (BoT) vectors that capture the similarity of texts within the same cluster. It consists of two stages. First, we select representative texts, which are then used to construct a BoT vector for each text. Second, we iteratively update each text's BoT vector by considering its similar texts until convergence. 

\subsection{Initial stage: construction of BoT representation}


\begin{algorithm}[t]
\small   
\caption{Select Representative Texts}
\label{alg:stage_cluster}
\begin{algorithmic}[1]

\REQUIRE $D$, $\mathbf{X}$, representative texts size $d$

\ENSURE $D^r$, $\mathbf{Z}^r \in \mathbb{R}^{d \times d}$

\STATE  $\mathcal{C} = \{ C_1, \cdots, C_d \} \gets \text{Agg}(\mathbf{X}, d), \quad C_j \subseteq \mathbf{X}$

\STATE $D^r \gets \emptyset$

\FOR{$j = 1, \dots, d$}
    \STATE $\mathbf{x}^r_j = \arg\min_{\mathbf{x} \in C_j} \sum_{\mathbf{y} \in C_j} \|\mathbf{x} - \mathbf{y}\|_2$
    \STATE $D^r \gets D^r \cup \{ x_i \in D : \mathbf{x}_i = \mathbf{x}^r_j \}$
\ENDFOR

\STATE Initialize $\mathbf{Z}^r \gets I_d$ \COMMENT{one-hot vectors for $D^r$}
\RETURN  Initial Representative Texts $D^{r}$, $\mathbf{Z}^r$

\end{algorithmic}
\end{algorithm}

\begin{algorithm}[t]
\caption{Construct Bag-of-Texts Vectors}
\small   
\label{alg:stage1}
\begin{algorithmic}[1]

\REQUIRE $D \setminus D^r$, $\mathbf{X}$, $D^r$, $\mathbf{Z}^r$, candidate size $m$, LLM

\ENSURE Bag-of-texts vectors $\mathbf{Z} \in \mathbb{R}^{N \times (d + d^*)}$

\STATE Initialize $\mathbf{z}_j \gets \mathbf{0} $ for all $x_j \in D \setminus D^r$

\FOR{each $x_j \in D \setminus D^r$}
    \STATE  $M_j\gets\{x_{j1},\dots,x_{jm}\} \in D^r$ from $\mathbf{X}$ as the top-$m$ by cosine similarity to $x_j$

    \STATE $O_j \gets \text{LLM}(M_j) \in \mathcal{P}(M_j) \cup \{\emptyset
\}$ \COMMENT{\# $\mathcal{P}(.)$: power set}

    \IF{$O_j \neq \emptyset
$}
        \STATE $\mathbf{z}_j \gets \text{MeanPooling}\{\mathbf{z}_k : x_k \in O_j\}$ 
        \STATE $\mathbf{z}_j \gets \frac{\mathbf{z}_j}{\|\mathbf{z}_j\|_2}$ 

    \ENDIF
\ENDFOR

\STATE $D^{r^*} \gets \{x_j: \mathbf{z}_j =\mathbf{0}\}$ \COMMENT{\# new reps. $|D^{r^*}| = d^*$}

\STATE $D^r \gets D^r \cup D^{r^*}$ \COMMENT{\# $|D^r| = d + d^*$}

\STATE Let $\mathbf{P} \gets \{\mathbf{z}_j : \mathbf{z}_j = \mathbf{0}\}$; one-hot encode each $\mathbf{z}_j \in \mathbf{P}$ in new dimensions, and pad the first $d$ columns with zeros.

\STATE Let $\mathbf{Q} \gets \{\mathbf{z}_j : \mathbf{z}_j \neq \mathbf{0}\}$; pad $\mathbf{Q}$ and $\mathbf{Z}^r$ with $d^*$ zero columns

\RETURN $\mathbf{Z} \gets \text{Stack}(\mathbf{Z}^r,\mathbf{P}, \mathbf{Q})$, rows ordered according to $D$

\end{algorithmic}
\end{algorithm}

We first encode all texts $D$ with a pre-trained embedder to obtain embeddings $\mathbf{X}$. To select initial representative texts $D^r$, we perform agglomerative clustering \citep{murtagh2014ward} to partition them into $d$ clusters, $d$ being a trivially large number (see Section~\ref{sec:hyperparam}). The medoid of each cluster is chosen as a representative and converted into a one-hot vector with initially equal spacing. See Algorithm~\ref{alg:stage_cluster} for specification.

The remaining texts are positioned based on LLM preference. Specifically, for each text $x_j \in D \setminus D^r$, we retrieve the top $m$ representative texts from the pretrained embeddings $\mathbf{X}$ that are closest based on cosine similarity. An LLM is then used to select which representative(s) are similar. The bag-of-texts vector for each text is computed by averaging the vectors of its selected representatives. After processing all texts, those with no selected representatives are promoted to new representatives and added to $D^r$ (so that $|D^r| = d + d^*$, where $d^*$ is the number of such new representatives), and the dimensionality of the BoT vectors is expanded to match. See Algorithm~\ref{alg:stage1} for specification.

This procedure ensures that the initial BoT vectors are evenly spaced for the representative texts, while the vectors for the remaining texts encode LLM-informed relationships, avoiding the need to use the full dimensionality of all texts.


\subsection{Iterative stage: bag-of-texts vectors refinements}

After constructing the initial positions for all texts, we iteratively update the BoT vectors to better align their relative distances with LLM preferences. The update procedure follows the same approach as in Algorithm~\ref{alg:stage1}, with three key modifications. First, the update uses all texts as the pool for subset selection, since each text has a BoT vector. Second, for each text, we select a subset of $m$ candidates based on the concatenation of its pre-trained embedding and BoT vector, $\mathbf{U}^t \triangleq \mathbf{X}_j \oplus \mathbf{Z}_j^t$. This ensures that the pre-trained embedding initially guides the BoT vectors, which carry minimal information at the start. Third, the update is performed iteratively rather than just once, allowing the candidate set to vary dynamically as the BoT vectors are refined. The update for a text stops when its BoT vector changes negligibly, as measured by a cosine similarity greater than 0.99. We set $m = 30$ and cap the maximum number of iterations at $T = 10$ as budget. See Algorithm~\ref{alg:stage2} in Appendix~\ref{app:stage2} for specification. Note that the pretrained embeddings $\mathbf{X}$ are used only for updating the Bot vectors; only the final BoT vectors $\mathbf{Z}^T$ will be used for clustering.

\section{Experimental Settings}

\begin{table}[t]
\centering
\small{
\begin{tabular}{l|c|c|c}
\hline
\textbf{Emerging data scenario} & $K$ & $|D^{test}|$ & $|D^u|$ \\ 
\hline
Bank77     & 77     & 3,080  & -    \\ 
Clinc150    &150     & 4,500   & -    \\ 
Mtop & 102     & 4,386   & -   \\ 
Massive   & 59     & 2,974  & -    \\ 
GoEmo   & 27     & 5,940   & -   \\ 
\hline
\textbf{GCD scenario}  & $K$ & \ $|D^{test}|$ & $|D^u|$\\ 
\hline

Bank77     & 77     & 3,080 & 2,700  \\ 
Clinc150    &150     & 2,250 & 5,400  \\ 
Stackoverflow& 20     & 1,000 & 5,400  \\ 
\hline

\end{tabular}
}
\caption{Dataset Statistics in two scenarios. \textbf{Emerging data scenario}: we use the same settings as \citet{zhang2023clusterllm, lin2025spill}. \textbf{GCD scenario}: for fair comparison, we evaluate on the same test sets as \citet{de2023idas,liang2024synergizing,zou2025glean}, but we only use $30\%$ of the original unlabeled training set. Further details are provided in Appendix \ref{GDC_data}.} 
\label{tab:dataset}
\end{table}

\subsection{Datasets} 
We use a variety of datasets to benchmark our method: the intent datasets Bank77 \citep{Casanueva2020}, CLINC150 \citep{larson2019evaluation}, Mtop \citep{li2020mtop}, and Massive \citep{fitzgerald2023massive}; the emotion dataset GoEmo \citep{demszky2020goemotions}, and the community QA dataset Stackoverflow \citep{xu2015short}. Statistics are provided in Table \ref{tab:dataset}. \update{Among these, MTOP, Massive, and GoEmo have highly imbalanced clusters.}

\subsection{Models} 
To see how sensitive our approach is to the choice of embedder, we design experiments that incorporate multiple models. We experiment in the \textbf{emerging data scenario} using TF–IDF, bert-base-uncased \citep{DBLP:journals/corr/abs-1810-04805}, all-MiniLM-L6-v2 \citep{reimers-2019-sentence-bert}, and e5-large \citep{wang2022text}. To see how far our method is from semi-supervised approaches and the effect with additional data, in the \textbf{GCD scenario}, we use all-MiniLM-L6-v2 (22.7M parameters) as the embedder following prior work \citep{rodriguez2024intentgpt}. For the LLM experiments, the main results are obtained using gemma-2-9b-it \citep{riviere2024gemma}. An A100 GPU or an H100 GPU were used throughout the experiments.

\paragraph{Prompts used}

The prompts we used for the LLM are shown in Appendix \ref{llm_prompt}. Following prior work \citep{zhang2023clusterllm, de2023idas,viswanathan-etal-2024-large, lin2025spill, zou2025glean}, we keep the prompts simple, as is standard practice, to ensure a fair comparison and generalizability between models.

\subsection{Hyperparameter setting} \label{sec:hyperparam}
The main hyperparameter in our approach is the initial dimension $d$ (number of initial representatives). Although we can simply set the number of $d$ equal to the total number of examples $N$, this becomes memory-intensive when $N$ is extremely large. Following the approach in \citep{feng2024llmedgerefine, lin2025spill}, we determine hyperparameters using external datasets to avoid biasing evaluation on the test sets. The search space starts from 512, setting a sufficiently large minimum to ensure that $d$ to include at least one representative from each cluster. We fix the BoT vector dimension to 1,024, based on validation results from two external datasets, including ArxivS2S and Reddit \citep{muennighoff2023mteb}. Empirically, we find that performance remains very similar between different values of $d$ (see Appendix \ref{d_select}).

\subsection{Evaluation}



The Bag-of-Text vectors are evaluated under K-means.
K-means clustering is a widely adopted evaluation practice in short text clustering \cite{ zhang2023clusterllm, de2023idas,viswanathan2024large, liang2024actively, lin2025spill, zou2025glean}. Following convention, the number of clusters for K-means is set to match the number of ground-truth cluster number $K$ during evaluation. K-means is run over 5 different seeds and averaged. After we derive the clusters, we apply standard clustering metrics for evaluation. These metrics include normalized mutual information (NMI) and clustering accuracy (Acc) \citep{rand1971objective, meilua2007comparing, huang2014deep, gung2023intent}.

\subsection{Baselines}
We compare our method with SOTA representation learning baselines for text clustering. Table \ref{tab:baselines} summarizes these methods. For the \textbf{emerging data scenario}, we compare against IDAS \cite{de2023idas}, ClusterLLM \citep{zhang2023clusterllm}, and SPILL \citep{lin2025spill}. For the \textbf{GCD Scenario} we compare against IntentGPT \cite{rodriguez2024intentgpt}. We also include comparisons with methods that require some labeled data, including LOOP \cite{an2024generalized}, ALUP \cite{liang2024actively} and Glean \cite{zou2025glean} to measure how far our method is from the performance of these semi-supervised approaches.

\begin{table}[t]
\centering
\resizebox{0.50\textwidth}{!}{
\begin{tabular}{l|p{0.8cm}|p{1cm}|p{0.6cm}|p{1cm}}
\hline
& {\footnotesize\textbf{Unsup}} & {\footnotesize\textbf{LLM}}&  {\footnotesize\textbf{FT}} &  {\footnotesize$\mathbf{K}$}\\ 
\hline
\textbf{LeBoT (Ours)}&$\checkmark$ &{\footnotesize Gemma}      &        &      \\
IDAS \cite{de2023idas} &$\checkmark$    & {\footnotesize Gemma}   &$\checkmark$  & $\checkmark^*$ \\ 

ClusterLLM \cite{zhang2023clusterllm}  &$\checkmark$& {\footnotesize GPT-*}    &     $\checkmark$ &  $\checkmark$    \\ 
IntentGPT \cite{rodriguez2024intentgpt} &$\checkmark^*$& {\footnotesize GPT-*}    &    &  \\ 
ALUP \citep{liang2024actively} &   & {\footnotesize GPT-*}  &   $\checkmark$ &  $\checkmark$\\ 
LOOP \cite{an2024generalized} &   & {\footnotesize GPT-*} &   $\checkmark$ &  $\checkmark$\\ 
SPILL \cite{lin2025spill} &$\checkmark$   & {\footnotesize Gemma}      &      \\ 
Glean \cite{zou2025glean} &   & {\footnotesize GPT-*} &   $\checkmark$ &  $\checkmark$\\ 
\hline
\end{tabular}
}

\caption{LLM-based representation learning for short-text clustering. \textbf{Unsup}: No access to labels. IntentGPT can be semi-supervised as well. \textbf{LLM}: Model used in the main results (Gemma = Gemma2-9b-it; GPT-* = GPT variants with unknown parameters). \textbf{FT}: Finetuning. $\mathbf{K}$: Incorporates the number of clusters $K$ into embedding refinement. IDAS can be applied either without knowing $K$ or knowing $K$}
\label{tab:baselines}
\end{table}


\section{Results}

\subsection{Main results}
\paragraph{Emerging data scenario} 

Table \ref{tab:EmgK} reports the K-means results. Our approach consistently outperforms pretrained embeddings and on average performs better than all other baselines, including those that require fine-tuning (ClusterLLM). Furthermore, our approach maintains consistent performance across backbones, unlike SPILL and IDAS, which degrade more when the embedder changes. \update{We also observe that with the stronger E5 embedder, which has the highest zero-shot performance, the gap between methods narrows. This is likely because E5 provides richer semantic representations and better separation in the embedding space, making it easier for both text-enrichment approaches (IDAS) and embedding post-processing methods (SPILL) to perform well. See Appendix \ref{embed_analysis} for detailed embedder analysis.}

\begin{table*}[t]
  \centering\
  \setlength{\tabcolsep}{2.5pt}
  \footnotesize
  \resizebox{\textwidth}{!}{
    \begin{tabular}{l|cc|lc|cc|cc|cc||cc}
    \hline
    &\multicolumn{2}{c|}{Bank 77} & \multicolumn{2}{c|}{Clinc150}  & \multicolumn{2}{c|}{Mtop} & \multicolumn{2}{c|}{Massive}& \multicolumn{2}{c||}{GoEmo} & \multicolumn{2}{c}{\textbf{Average}} \\
     \cline{1-3} \cline{4-5} \cline{6-7} \cline{8-9} \cline{10-11} \cline{12-13}   
    & NMI & Acc & NMI & Acc & NMI & Acc& NMI & Acc & NMI & Acc & NMI & Acc\\
    \hline

    \textbf{TF-IDF}&&&&&&&&&&&\\
    
    Plain & 59.76 \tiny{(0.31)} & 36.73 \tiny{(0.99)} & 58.05 \tiny{(0.43)} & 33.44 \tiny{(1.11)} & 50.77 \tiny{(0.84)} & 30.98 \tiny{(1.19)} & 45.24 \tiny{(1.55)} & 28.35 \tiny{(1.61)} & 18.49 \tiny{(1.65)} & 19.64 \tiny{(1.25)} & 46.46 & 29.83 \\

    IDAS & 72.62 \tiny{(0.28)} & 51.96 \tiny{(0.78)} & 76.37 \tiny{(0.68)} & 54.44 \tiny{(1.04)} & 63.95 \tiny{(0.11)} & 39.47 \tiny{(2.90)} & 58.07 \tiny{(1.08)} & 40.82 \tiny{(1.27)} & 19.49 \tiny{(0.18)} & 21.04 \tiny{(0.47)} & 58.10 & 41.55 \\

     IDAS$^\dagger$ & 72.68 \tiny{(0.20)} & 52.11 \tiny{(0.60)} & 85.12 \tiny{(0.14)} & 69.72 \tiny{(0.29)} & 65.58 \tiny{(0.24)} & 38.33 \tiny{(0.45)} & 65.14 \tiny{(0.32)} & 46.77 \tiny{(0.85)} & 23.39 \tiny{(0.36)} & 22.48 \tiny{(0.45)} & 62.38 & 45.88 \\

    SPILL & 63.73 \tiny{(0.67)} & 42.36 \tiny{(1.14)} & 59.72 \tiny{(1.05)} & 36.74 \tiny{(0.92)} & 49.54 \tiny{(1.41)} & 28.26 \tiny{(0.94)} & 43.26 \tiny{(0.85)} & 28.06 \tiny{(1.28)} & 9.51 \tiny{(0.44)} & 13.48 \tiny{(0.20)} & 45.15 & 29.78 \\

    LeBot (Ours)& \textbf{80.26} \tiny{(0.28)} & \textbf{65.45} \tiny{(1.43)} & \textbf{88.97} \tiny{(0.11)} & \textbf{77.70} \tiny{(1.03)} & \textbf{70.15} \tiny{(0.23)} & \textbf{40.00} \tiny{(1.30)} & \textbf{69.60} \tiny{(0.06)} & \textbf{54.52} \tiny{(0.69)} & \textbf{27.89} \tiny{(0.30)} & \textbf{25.82} \tiny{(0.34)} & \textbf{67.37} & \textbf{52.70} \\
    \hline

    \textbf{Bert}&&&&&&&&&&&\\

    Plain & 50.72 \tiny{(0.42)} & 29.07 \tiny{(0.85)} & 72.90 \tiny{(0.60)} & 49.48 \tiny{(0.83)} & 62.24 \tiny{(0.46)} & 26.78 \tiny{(0.32)} & 49.12 \tiny{(0.64)} & 30.49 \tiny{(1.00)} & 10.77 \tiny{(0.46)} & 12.90 \tiny{(0.37)} & 49.15 & 29.75 \\

    IDAS &70.30 \tiny{(0.32)} & 49.85 \tiny{(0.49)} & 87.97 \tiny{(0.27)} & 71.57 \tiny{(1.02)} & 68.90 \tiny{(0.29)} & 33.08 \tiny{(0.69)} & 65.24 \tiny{(0.32)} & 49.06 \tiny{(0.59)} & 14.10 \tiny{(0.52)} & 15.22 \tiny{(0.73)} & 61.30 & 43.76 \\

    IDAS$^\dagger$ & 72.73 \tiny{(0.26)} & 53.45 \tiny{(0.82)} & 88.70 \tiny{(0.18)} & 74.99 \tiny{(0.61)} & 69.52 \tiny{(0.21)} & 33.19 \tiny{(0.87)} & 66.41 \tiny{(0.44)} & 50.04 \tiny{(0.94)} & 14.89 \tiny{(0.26)} & 16.50 \tiny{(0.52)} & 62.45 & 45.63 \\
    SPILL & 65.67 \tiny{(0.49)} & 43.44 \tiny{(1.02)} & 86.91 \tiny{(0.28)} & 71.39 \tiny{(1.28)} & 69.50 \tiny{(0.28)} & 32.49 \tiny{(0.40)} & 62.67 \tiny{(0.29)} & 43.64 \tiny{(0.81)} & 12.30 \tiny{(0.32)} & 13.86 \tiny{(0.44)} & 59.41 & 40.96 \\

    LeBoT (Ours) & \textbf{76.86} \tiny{(0.24)} & \textbf{60.95} \tiny{(1.11)} & \textbf{92.90} \tiny{(0.05)} & \textbf{84.47} \tiny{(0.46)} & \textbf{72.28} \tiny{(0.24)} & \textbf{38.69} \tiny{(1.46)} & \textbf{73.16} \tiny{(0.27)} & \textbf{61.02} \tiny{(0.82)} & \textbf{19.25} \tiny{(0.19)} & \textbf{18.77} \tiny{(0.14)} & \textbf{66.89} & \textbf{52.78} \\
    
    \hline

    \textbf{MiniLM}&&&&&&&&&&&\\

    Plain & 79.08 \tiny{(0.80)} & 61.27 \tiny{(1.36)} & 89.46 \tiny{(0.19)} & 73.73 \tiny{(0.95)} & 67.41 \tiny{(0.53)} & 31.40 \tiny{(1.62)} & 70.37 \tiny{(0.62)} & 54.04 \tiny{(1.28)} & 10.84 \tiny{(0.16)} & 13.89 \tiny{(0.63)} & 63.43 & 46.87 \\

    IDAS & 83.01 \tiny{(0.19)} & 65.96 \tiny{(1.19)} & 92.82 \tiny{(0.22)} & 80.05 \tiny{(0.21)} & 70.46 \tiny{(0.46)} & 35.55 \tiny{(1.33)} & 75.67 \tiny{(0.47)} & 59.52 \tiny{(2.12)} & \textbf{22.27} \tiny{(0.16)} & \textbf{27.22} \tiny{(0.37)} & 68.85 & 53.66 \\

    IDAS$^\dagger$ & 84.88 \tiny{(0.19)} & 72.77 \tiny{(0.59)} & 93.41 \tiny{(0.06)} & 85.52 \tiny{(0.33)} & 70.88 \tiny{(0.22)} & 36.46 \tiny{(1.46)} & 75.93 \tiny{(0.17)} & 57.75 \tiny{(0.57)} & 21.25 \tiny{(0.25)} & 25.67 \tiny{(0.45)} & 69.27 & 55.63 \\
    
    SPILL & 81.92 \tiny{(0.16)} & 66.22 \tiny{(0.46)} & 90.78 \tiny{(0.22)} & 77.76 \tiny{(0.07)} & 70.48 \tiny{(0.41)} & 37.95 \tiny{(1.27)} & 72.59 \tiny{(0.31)} & 58.05 \tiny{(1.40)} & 15.54 \tiny{(0.31)} & 17.22 \tiny{(0.75)} & 66.26 & 51.44 \\

    LeBoT (Ours) & \textbf{85.44} \tiny{(0.18)} & \textbf{73.12} \tiny{(1.10)} & \textbf{94.45} \tiny{(0.08)} & \textbf{87.58} \tiny{(0.70)} & \textbf{72.18} \tiny{(0.29)} & \textbf{39.40} \tiny{(0.83)} & \textbf{76.74} \tiny{(0.24)} & \textbf{66.84} \tiny{(0.67)} & 21.77 \tiny{(0.12)} & 25.22 \tiny{(0.74)} & \textbf{70.12} & \textbf{58.43} \\
    \hline
    \textbf{E5}&&&&&&&&&&&\\

    Plain  & 77.86 \tiny{(0.35)} & 60.88 \tiny{(0.89)} & 91.09 \tiny{(0.17)} & 75.70 \tiny{(0.36)} & 71.17 \tiny{(0.28)} & 33.04 \tiny{(0.74)} & 71.70 \tiny{(0.84)} & 53.97 \tiny{(2.00)} & 21.13 \tiny{(0.44)} & 21.93 \tiny{(0.60)} & 66.59 & 49.10 \\

    IDAS & 83.91 \tiny{(0.24)} & 68.78 \tiny{(1.11)} & 93.53 \tiny{(0.26)} & 82.20 \tiny{(1.22)} & 73.58 \tiny{(0.37)} & 38.87 \tiny{(1.04)} & 76.70 \tiny{(0.89)} & 59.61 \tiny{(3.25)} & 26.30 \tiny{(0.17)} & 29.56 \tiny{(0.64)} & 70.80 & 55.80 \\

    IDAS$^\dagger$ &  83.37 \tiny{(0.14)} & 70.01 \tiny{(0.59)} & \textbf{94.07} \tiny{(0.09)} & 85.38 \tiny{(0.35)} & 73.09 \tiny{(0.45)} & 39.84 \tiny{(1.24)} & \textbf{77.29} \tiny{(0.15)} & \textbf{63.64} \tiny{(0.80)} & 25.99 \tiny{(0.09)} & 28.72 \tiny{(1.08)} & 70.76 & 57.52 \\

    SPILL& 83.56 \tiny{(0.47)} & 70.25 \tiny{(1.56)} & 92.93 \tiny{(0.08)} & 83.18 \tiny{(0.78)} & 71.77 \tiny{(0.35)} & 36.83 \tiny{(1.10)} & 75.40 \tiny{(0.51)} & 60.28 \tiny{(1.81)}& 25.14 \tiny{(0.36)} & 24.82 \tiny{(0.86)}  &	69.76	&	55.07\\

    ClusterLLM$^\dagger$
   &  84.16 
    \tiny{(0.36)} &70.13 \tiny{(1.34)}    &  92.92 \tiny{(0.29)} &80.48 \tiny{(0.93)} & \textbf{74.46} \tiny{(0.11)} &37.22 \tiny{(1.18)}  &  74.39 \tiny{(0.21)} & 56.08 \tiny{(1.01)}& 22.23 \tiny{(0.17)} & 22.22 \tiny{(1.15)} &	69.63	&	53.23\\

    LeBoT (Ours) & \textbf{85.34} \tiny{(0.06)} & \textbf{73.86} \tiny{(0.76)} & 93.63 \tiny{(0.06)} & \textbf{85.65} \tiny{(0.46)} & 72.51 \tiny{(0.32)} & \textbf{40.53} \tiny{(1.64)} & 76.16 \tiny{(0.17)} & 62.75 \tiny{(0.67)} & \textbf{26.70} \tiny{(0.04)} & \textbf{33.70} \tiny{(0.26)} & \textbf{70.87} & \textbf{59.30} \\

    \hline

  \end{tabular}
  }

  \caption{\update{Five-benchmark results (\textbf{emerging data scenario}). Averages over 5 runs (±SD). Gemma is used in LeBoT, SPILL and IDAS. \textbf{Plain}: direct embedding clustering. Bold numbers: best within the embedder. ClusterLLM cited directly from prior work. $^\dagger$ is with access to information of known $K$ during representation refinements.}}

  \label{tab:EmgK}
\end{table*}

\paragraph{GCD scenario}

Table \ref{tab:GCDK} shows the results for GCD. The goal is to examine how increasing the amount of data affects performance and to see the gap between our approach and semi-supervised learning. Our method achieves the best performance on all benchmarks in the fully unlabeled setting. It even outperforms baselines that know $10\%$ of the categories, and is comparable to other baselines that know $25\%$ of the categories (or up to $75\%$ in the case of IntentGPT). These strong results suggest that our approach can substantially reduce the need for human labeling.

\begin{table}[t]
  \centering
  \setlength{\tabcolsep}{2.5pt}
    \resizebox{0.48\textwidth}{!}{
    \begin{tabular}{l|l|cc|cc|cc|cc}
    \hline
     & &  \multicolumn{2}{c|}{Bank 77} & \multicolumn{2}{c|}{Clinc150}  & \multicolumn{2}{c|}{StackO}& \multicolumn{2}{c}{Average} \\ \cline{3-4}   
     \cline{5-6} \cline{7-8} \cline{9-10}  
    KCR & Method & NMI & Acc & NMI & Acc & NMI & Acc & NMI & Acc \\

    \hline
    $\mathbf{0\%}$&&&&&&&&\\

    & IntentGPT & 81.42 & 64.22 &  94.35 &  83.20 &-	&- & - & -\\

    & LeBoT (Ours) & \textbf{86.90} & 74.93 & 95.32 & 88.71 & \textbf{81.21} & \underline{82.54} & \textbf{87.81} & \underline{82.06} \\
    \hline
    $\mathbf{10\%}$&&&&&&&&\\
   & LOOP$^\dagger$ & 79.14 & 64.97 & 93.52 & 84.89 & 75.98 & 80.50 & 82.88 & 76.79 \\
    & GLEAN$^\dagger$ & 82.23 & 67.99 & 95.21 & 88.71 & 79.67 & 82.40 &	85.70	&	79.70\\
    \hline
    $\mathbf{25\%}$&&&&&&&&&\\

   & ALUP$^\dagger$ & 84.06 & 74.61 & 94.84 & 88.40 & 76.58 & 82.20 & 85.16 & 81.74\\
  & LOOP$^\dagger$ & 83.37 & 71.40 & 94.38 & 86.58 & 79.10 & 82.20 & 85.62 & 80.06 \\
    & GLEAN$^\dagger$& 85.62 & \underline{76.98} & \textbf{96.27} & \textbf{91.51} & \underline{80.90} & \textbf{84.10} &	87.60	&	\textbf{84.20}\\
    \hline 
    $\mathbf{75\%}$& IntentGPT &\underline{85.94}	& \textbf{77.21} &\underline{96.06}	&\underline{88.76}	&-	&- & - & -\\ 
    \hline    
  \end{tabular}
  }

    \caption{Three-benchmark results (\textbf{GCD scenario}). Averages over 5 runs. KCR: Known Categories Ratio. All-MiniLM-L6-v2 used as in LeBoT. Bold: best. Underline, second best. StackO: StackOverflow dataset. GLEAN and ALUP cited directly from prior work. LOOP cited directly from GLEAN. $^\dagger$ is with access to information of known $K$ during representation refinements.}

  \label{tab:GCDK}
\end{table}

\subsection{Further analysis}

\begin{figure}[t]
        \centering
        \includegraphics[width=0.45\textwidth]{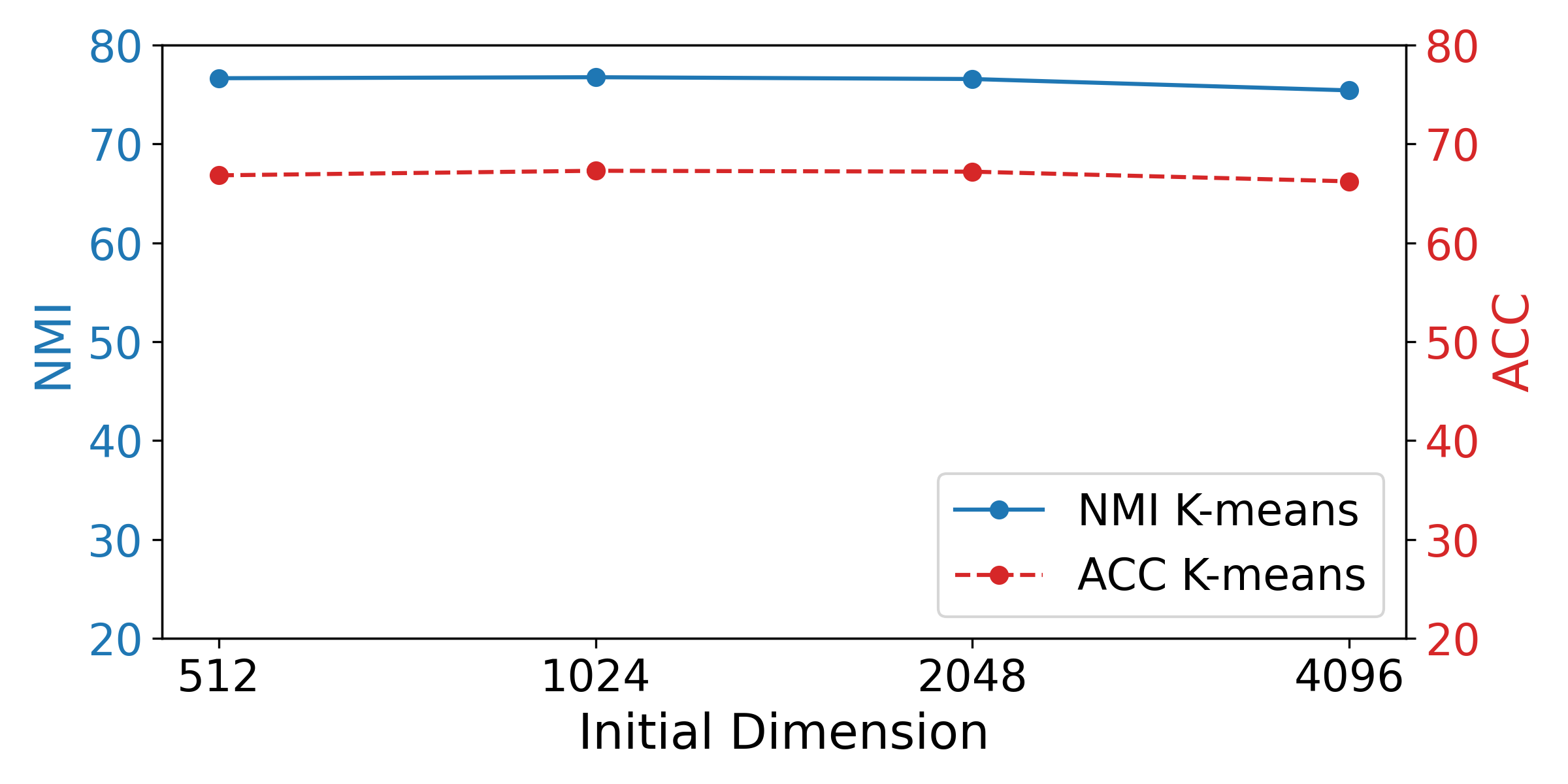}
        \caption{ACC and NMI across all test datasets (Emerging Data and GCD scenarios) for different $d$. $4,096$ means $\min\{4096, N\}$ since some datasets are smaller.}
        \label{fig:dim_test}
\end{figure}

\paragraph{Dimensionality of the bag-of-text representations}
We observe that the increase in dimensionality by $d^*$ is negligible: for most runs, the values fall between 0 and 20, and the maximum increase across all runs, datasets, and backbone embedders is only 44. Since we start with a relatively large initial dimension of $d = 1,024$, this increase is very limited. This is expected, as each text is more likely to identify similar counterparts when the pool of candidate texts is large. We also tested other initial dimensions, and Figure \ref{fig:dim_test} shows that the results remain robust across different initial settings.

\begin{figure}[t]
    \centering

    \begin{subfigure}[b]{0.39\textwidth}
        \centering
        \includegraphics[width=\textwidth]{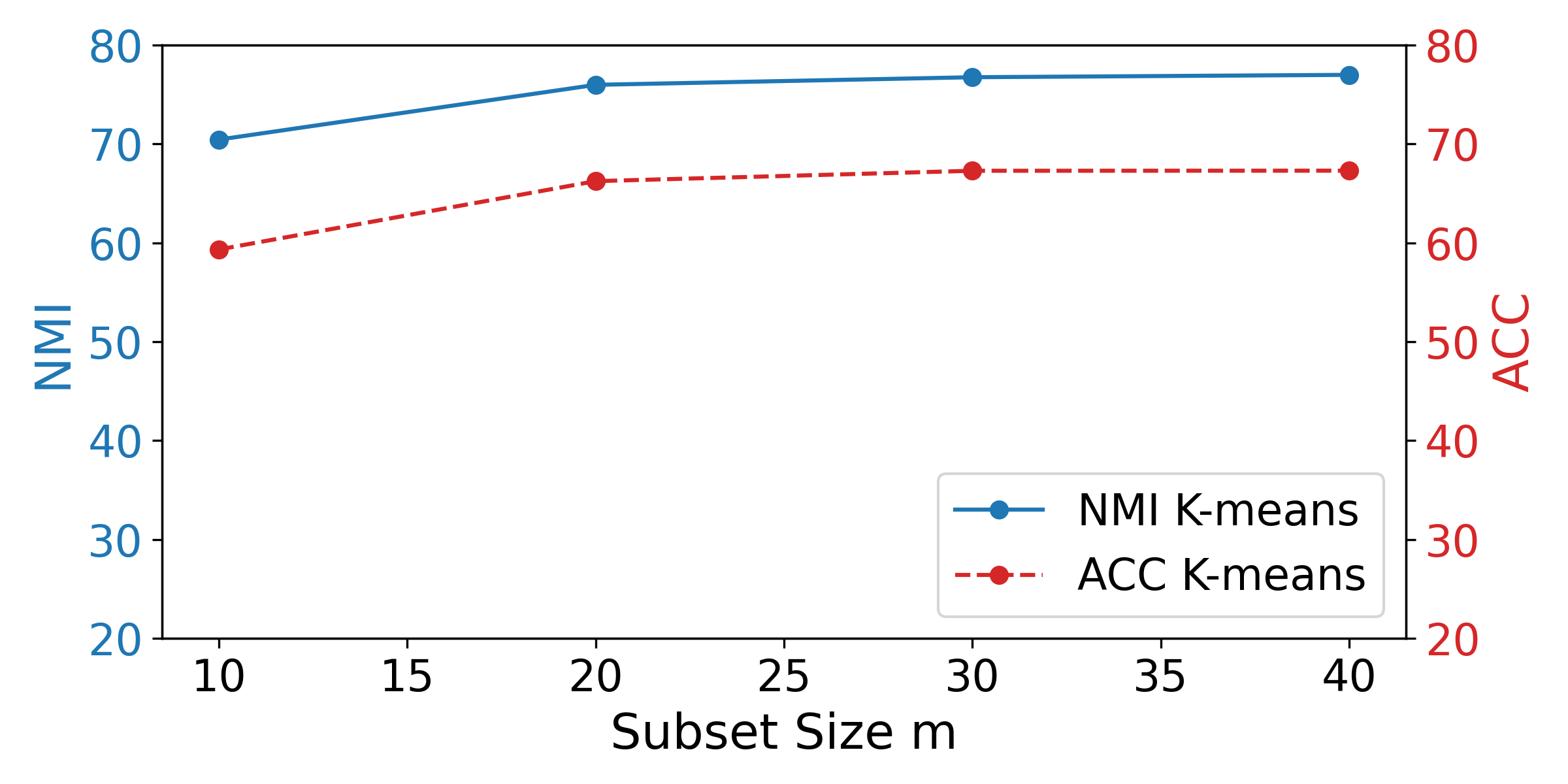}
        \caption{}
        \label{fig:Budget_sub}
    \end{subfigure}
    \hfill
    \begin{subfigure}[b]{0.39\textwidth}
        \centering
        \includegraphics[width=\textwidth]{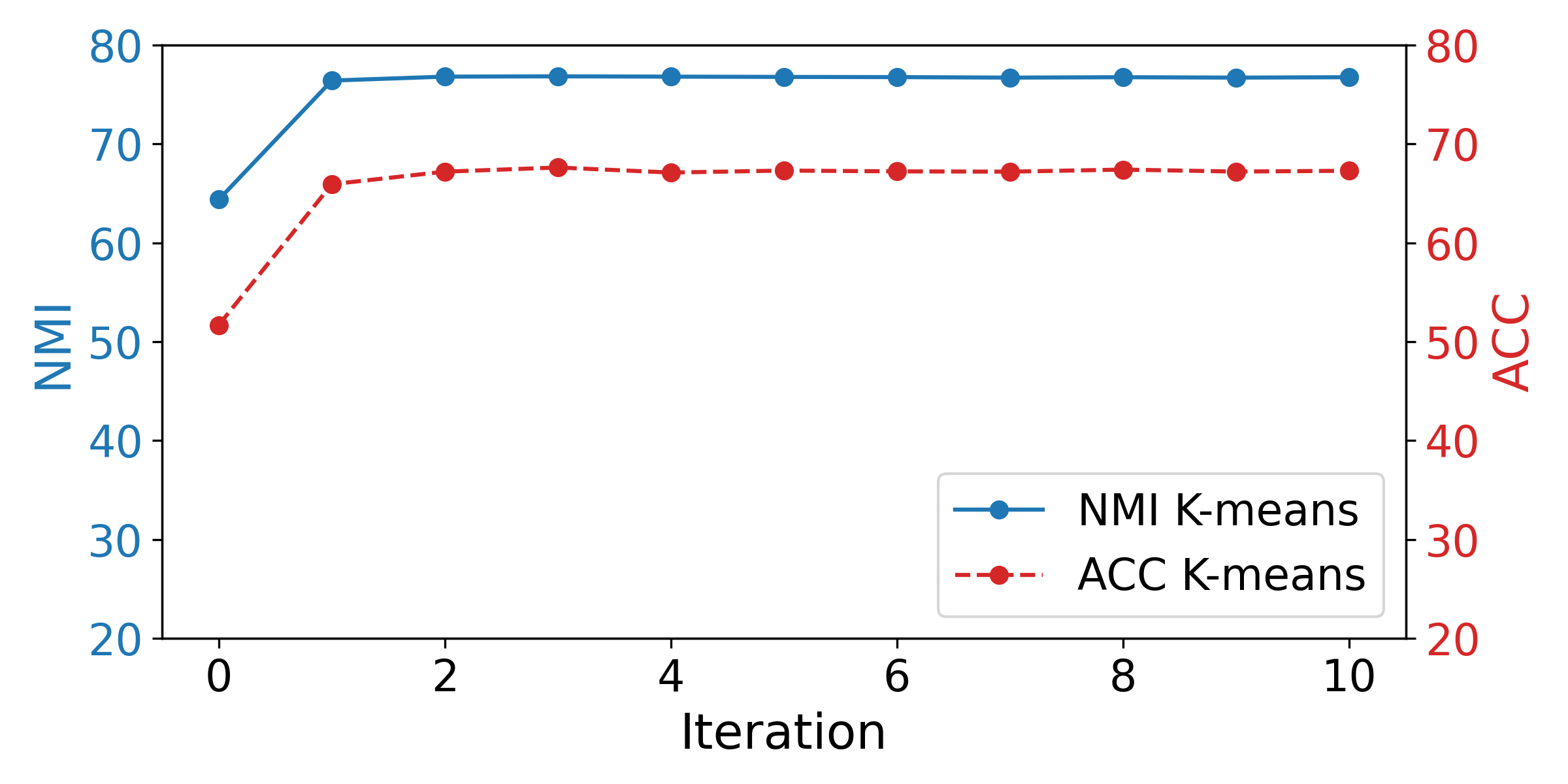}
        \caption{}
        \label{fig:Iter_sub}
    \end{subfigure}

    \caption{Comparison of (a) varying neighbor $m$ and (b) iterative updates on ACC and NMI across all test datasets (Emerging Data and GCD). Backbone embedder:  All-MiniLM-L6-v2.}
    \label{fig:combined}
\end{figure}

\paragraph{Effect of the Number of subset candidates \textbf{m} and Iterative Updates} We use $m = 30$ across all experiments, but this raises the question of whether different values of $m$ would affect performance. Figure \ref{fig:Budget_sub} shows that small size of $m = 10$ leads to lower results because the LLM has too few candidates to choose from. However, once $m \geq 20$, performance remains largely stable. To investigate the effect of iterative updates, Figure \ref{fig:Iter_sub} shows that in subsequent iterations, the results substantially improve, especially after the first iteration, but no longer after the  or third. Since most BoT vectors converge within 5 iterations, there are only minimal changes afterward: We find that most datasets have 90\% convergence ratio by the 5th iteration (see Appendix \ref{convergence}).

\paragraph{Effect of Different LLMs}

\begin{figure}[t]
        \centering
        \includegraphics[width=0.39\textwidth]{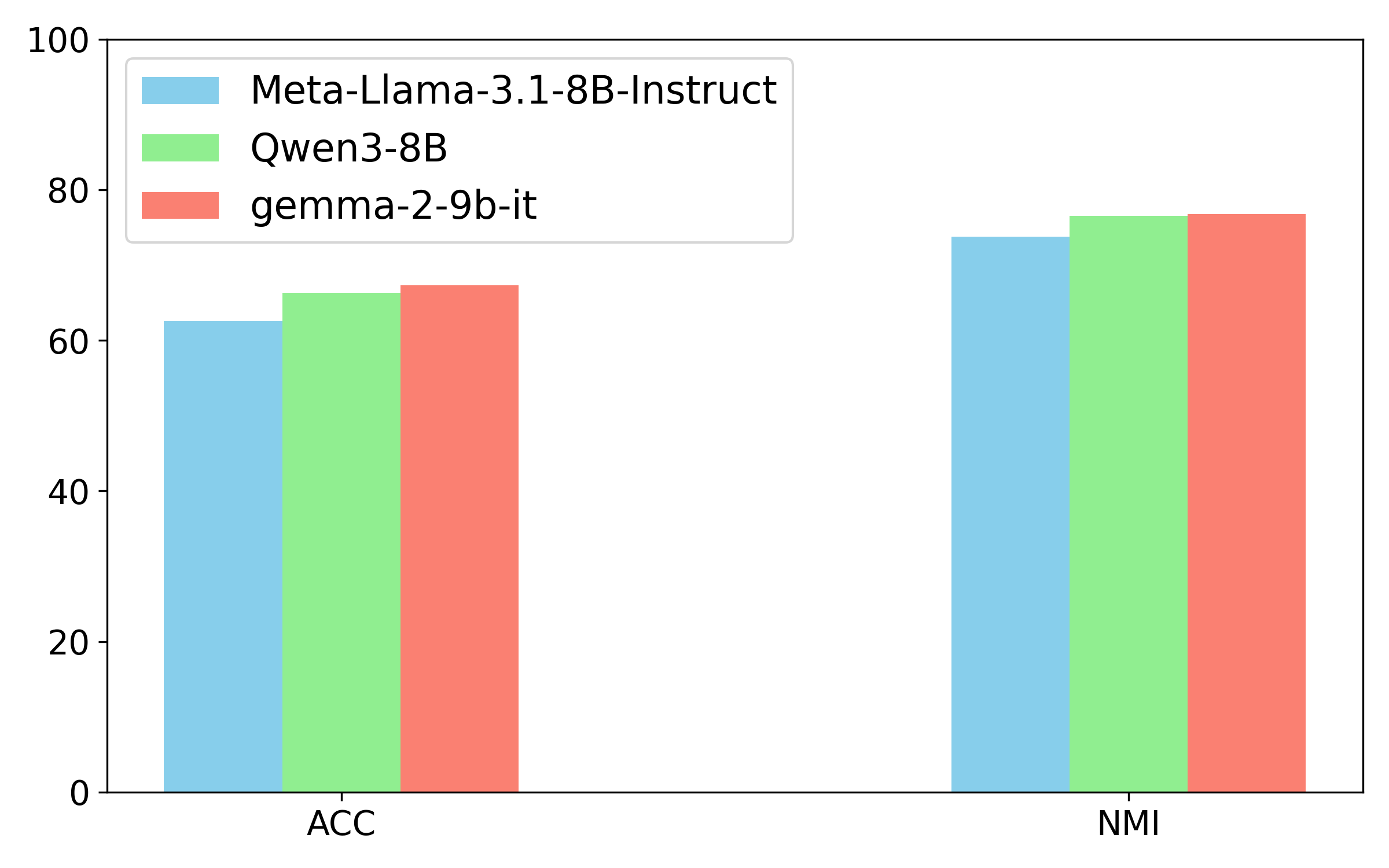}
        \caption{ACC and NMI across all test datasets (Emerging Data and GCD) for three LLMs. Backbone embedder: All-MiniLM-L6-v2}
        \label{fig:LLM_result}
\end{figure}

Our main results were obtained with gemma-2-9b-it. For completeness, we also experimented with alternative LLMs. \update{Importantly, because our objective is to evaluate whether the vectors encode LLM-derived selections rather than to benchmark LLM performance, these analyses are presented as ancillary robustness checks.} Figure \ref{fig:LLM_result} shows the results obtained with three different LLMs. The results between Qwen and Gemma are similar. There is a slight drop with Llama. This finding is consistent with prior work \cite{lin2025spill}, as Llama produces more incorrect selections (see Appendix \ref{LLM_result_by_dataset} for details).

\paragraph{Evaluation with different clustering algorithms}

\begin{figure}[t]
        \centering
        \includegraphics[width=0.39\textwidth]{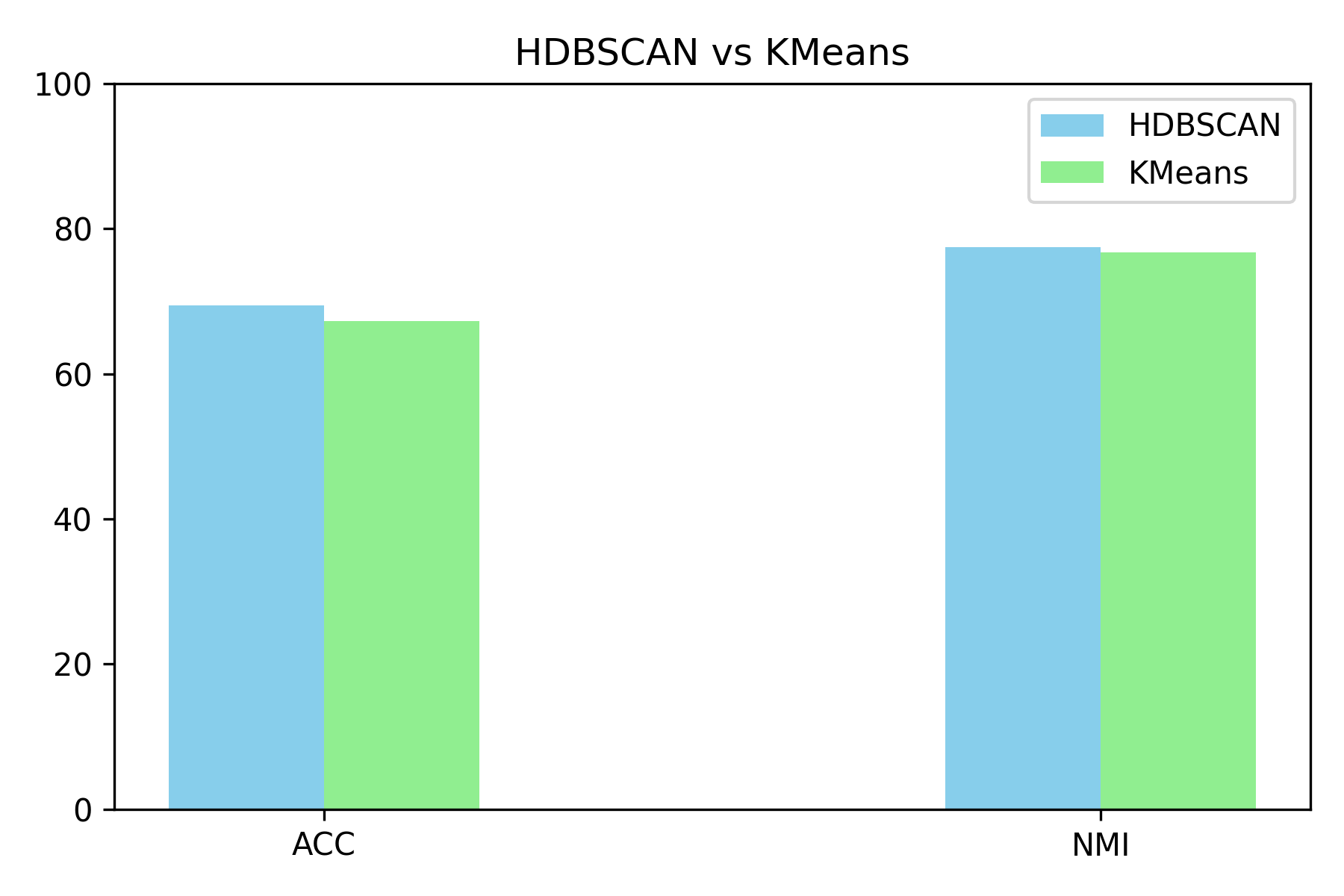}
        \caption{ACC and NMI across all test datasets (Emerging Data and GCD) for HDBSCAN and K-means. Backbone embedder: All-MiniLM-L6-v2}
        \label{fig:HDBSCAN_result}
\end{figure}

Our approach does not require the number of clusters $K$ when building the representations. We therefore also evaluate it with HDBSCAN, which estimates the number of clusters automatically and is therefore closer to real-world use. Figure \ref{fig:HDBSCAN_result} shows that performance remains similar, indicating that our method is robust across different clustering algorithms. This confirms its practical applicability. Detailed implementation in Appendix \ref{HDBSCAN}.


\paragraph{Increasing scalability with distillation} Our representation requires on average 3 iterations to reach stable results, which can be computationally expensive when the dataset is extremely large. To assess scalability, we additionally explore clustering over the full CLINC (22,500 examples) and Banking (13,083 examples) datasets. The goal is to implement our approach on a subset of the texts (around 20$\%$) and then use a model to generalize to the remaining texts (a form of knowledge distillation), without requiring further LLM selection. Specifically, we train a lightweight MLP ($\sim$10M parameters) to map pre-trained embeddings to BoT vectors. See Appendix \ref{simu_dim} for experiment details. Table \ref{tab:GCDK_scale} shows that our approach still improves the original embeddings by a large margin. \update{During inference, our method with LLM selection took $0.7000 s$ (CLINC) and $0.7083 s$ (Banking) per text, while our distilled method with MLP only takes $0.0002 s$ per text for both.}

\section{Conclusions}
We propose LeBoT (LLM-enabled Bag-of-Texts), an intuitive, domain-adaptive, label-free, and training-free method for short text clustering. \update{Compared with other SOTA methods, LeBoT is less dependent on careful selection of the embedder and does not assume prior knowledge of clusters or labels, making it more aligned with real-world scenarios.} 
LeBoT achieves better or similar results to state-of-the-art approaches, including methods that require labeled data. \update{Experiments on diverse datasets shows our method is the most robust across different embedders}, and experiments with smaller LLMs show that our method is robust in low-resource settings, while also scalable to large datasets. 
In the future, we plan to incorporate human feedback into our method pipeline.

\begin{table}[t]
  \centering
  \setlength{\tabcolsep}{2.5pt}
    \resizebox{0.35\textwidth}{!}{
    \begin{tabular}{l|cc|cc}
    \hline
     &  \multicolumn{2}{c|}{Bank 77} & \multicolumn{2}{c}{Clinc150}  \\ \cline{2-3}   
     \cline{4-5} 
    Method & NMI & Acc & NMI & Acc \\

    \hline

    \textbf{K-means}&&&&\\
    Plain & 76.31 & 59.41 & 88.99 & 73.99   \\
    LeBoT (Ours) & 81.44 & 68.99 & 91.19 & 83.62  \\
    \hline
    \textbf{HDBSCAN}&&&&\\

    Plain & 76.15 & 60.27 & 84.31 & 62.84 \\  
    LeBoT (Ours)  &81.35 & 68.01 & 91.50 & 85.28\\
    \hline

  \end{tabular}
  }

    \caption{Clustering results on the entire large datasets using knowledge distillation. Averages over 5 runs for K-means. Plain: Directly embedding the test dataset with All-MiniLM-L6-v2.}

  \label{tab:GCDK_scale}
\end{table}
\clearpage

\section*{Acknowledgments}
This publication is part of the project LESSEN with project number NWA.1389.20.183 of the research program NWA ORC 2020/21 which is (partly) financed by the Dutch Research Council (NWO).

\section*{Limitations}

\textbf{Language.} Like most of the prior work, we only focus on English utterance datasets. This is relevant because most of benchmark datasets are in English. Our method is completely data-driven and not dependent on labeled data and it is therefore applicable to other languages than English.\\
\textbf{LLM capability.} Our method focuses on encoding preferences through Bag-of-Text vectors. Its effectiveness naturally depends on the LLM, and in highly specialized domains, some in-context learning may be required to improve the selection. \\
\textbf{Embedder capability.} Although we construct a Bag-of-Text vector that reduces reliance on the existing embedder, for computational efficiency we still retrieve subset of candidate texts using the embedder. This will affect the quality of the constructed vectors if the embedder is very poor. This can affect the quality of the constructed vectors when the embedder is very weak. However, our main results show that this influence is relatively small compared with other methods.
\noindent

\noindent

\bibliography{anthology,custom}
\bibliographystyle{acl_natbib}
\clearpage

\appendix

\section{\update{Impact of our computational method on preference preservation}}\label{rig_analysis}

In this section, we analyze the impact of dimension reduction and subset selection on the preservation of preference, assuming that the stated Assumption \ref{assum_partition_cluster} hold. We do not discuss pooling here, because pooling is intended to handle cases where Assumption \ref{assum_partition_cluster} is violated, and to account for the estimation involved in subset selection.

\begin{definition}[Preservation of Preference]
We say that a function \emph{preserves preferences} if $l_{x, x^{+}} > l_{x, x^{-}}$, where $l$ denotes the cosine similarity, $(x, x^{+})$ is a similar pair, and  $(x, x^{-})$ is a disimilar pair.
\end{definition}

We need a generalized version of $g$ to consider the dimension reduction and subset selection.

\begin{definition}[Bag-of-Texts: generalization version of $g$]
Let $g^*(x, \mathcal{T}, \mathcal{L}(x))$ be a function that takes:
\begin{itemize}
    \item a text $x$,
    \item a set of texts $D$,
    \item a subset of texts $\mathcal{T} \subseteq D$,
    \item a subset of texts $\mathcal{L}(x) \subseteq \mathcal{T}$, selected based on $x$ 
\end{itemize}
and outputs a BoT representation $\mathbf{z} \in \mathbb{R}^{|\mathcal{T}|}$. In this representation, the entry corresponding to each text in $\mathcal{L}(x)$ that is considered similar to $x$ is assigned a value of 1, while all other entries are 0.

\end{definition}

The definition of the generalized function $g^*$ is to consider the reduced dimension of BoT (memory constraints) and subset selection (computation constraints). Note $g^*(x,D,D,D)$ = $g(x,D)$.

\begin{proposition}[Preservation of Preference on generalization function $g^*$]\label{gen_g_prep}

Let $D^r \subseteq D$ be such that for every cluster $S \in \mathcal{S}$, there exists at least one text $y \in D^r$ with $y \in S$. 

Let $M^*= \mathcal{L}(x) \subseteq \mathcal{T}$ such that, for every cluster $S \in \mathcal{S}$, there exists at least one text $o_S \in S$ that is included in $M^*$ for $x \in S$.

$g^*(x,D, D^r,M^*)$ is a function that preserves the preference under Assumption 1.
\end{proposition}

In the preposition \ref{gen_g_prep}, the dimensions constructed using $D^r$ ensure that each cluster has at least one dimension in which a value can be assigned. The subset $M^*$ guarantees that BoT vectors corresponding to texts within the same cluster share a 1 in at least one of these dimensions. As a result, the cosine similarity between vectors of texts from the same cluster is strictly positive due to the shared dimension(s). Furthermore, the more shared texts from the cluster that appear in $M^*$ for each $x$, the greater the overlap of 1s, and thus the higher the cosine similarity. 

This proposition \ref{gen_g_prep} implies two key insights. First, instead of turning every text $ x \in D$ into a dimension of the BoT vector, we may use only a subset, provided that the subset is sufficiently representative. Second, scanning the entire dataset for every $x$ can be replaced with scanning only such a subset, as long as it contains at least one shared element for each cluster.

\textbf{Remark 1:} The proposition \ref{gen_g_prep} is primarily intended for conceptual understanding. The shared text $o_S$ and subset of texts $M^*_j$ can also include the texts from remaining texts $D \setminus D^r$ not just those in $D^r$ since these remaining texts are initially constructed with $D^r$. This increases the number of shared texts available across clusters, providing more opportunities for matching and potentially improving similarity estimates.

\textbf{Remark 2:} 
As mentioned in the section 3 and 4, in practice, we select a fairly large initial dimension $d$ and use an LLM to include missing representative texts to form $D^r$. For subset selection, for each $x_j$, we use pre-trained embeddings to select the top $m$ closest texts, denoted $M_j$, to approximate $M^*_j$.

\section{Algorithm for Updating the Bag-of-Texts Vectors} \label{app:stage2}

The BoT vector update largely involves a repetitive process similar to the original construction. See Algorithm \ref{alg:stage2} for details.

\begin{algorithm}[t]
\small 
\caption{Update Bag-of-Texts Vectors}

\label{alg:stage2}
\begin{algorithmic}[1]

\REQUIRE $D$, $\mathbf{X}$, $\mathbf{Z}$, candidates size $m$, max iterations $T$, LLM

\ENSURE Bag-of-texts vectors $\mathbf{Z} \in \mathbb{R}^{N \times (d + d^*)}$

\STATE Initialize $\mathbf{Z}^0 \gets \mathbf{Z}$

\FOR{$t = 0$ to $T-1$}
    \STATE $\mathbf{U}^t \gets \mathbf{X} \oplus \mathbf{Z}^t$
    \FOR{each $x_j \in D$}
        \IF{$t >0$ \AND $\cos(\mathbf{z}_j^{t-1}, \mathbf{z}_j^{t}) > 0.99$} 
        \STATE $\mathbf{z}_j^{t+1} \gets \mathbf{z}_j^t$ \COMMENT{\# Skip updating}

        \ELSE

            \STATE $M_j^t \gets\{x^t_{j1},\dots,x^t_{jm}\}$ from $\mathbf{U}^t$ as the top-$m$ by cosine similarity to $x_j$
            \STATE $O^{t}_j \gets \text{LLM}(M_j^t) \in \mathcal{P}(M_j) \cup \{\emptyset
\}$ \COMMENT{\# $\mathcal{P}(.)$: power set}
        
            \IF{$O^t_j \neq\emptyset
$}
        \STATE $\mathbf{z}^{t+1}_j \gets \text{MeanPooling}\{\mathbf{z}^t_k : x_k \in O^t_j\}$ 
        \STATE $\mathbf{z}^{t+1}_j \gets \frac{\mathbf{z}^{t+1}_j}{\|\mathbf{z}^{t+1}_j\|_2}$ 
    
            \ELSE
                \STATE $\mathbf{z}_j^{t+1} \gets \mathbf{z}_j^t$
            \ENDIF
        \ENDIF
    \ENDFOR
    \STATE $\mathbf{Z}^{t+1} \gets \{\mathbf{z}_j^{t+1}\}_{x_j \in D}$

\ENDFOR
\RETURN $\mathbf{Z}^T  \in \mathbb{R}^{N \times (d + d^*)}$

\end{algorithmic}
\end{algorithm}

\section{Comparison of Data Splits in GCD scenario: Our Method vs. Prior Work}\label{GDC_data}

Table \ref{tab:GCDSplit} shows the train–validation–test splits used in ALUP \citep{liang2024actively} and GLEAN \cite{zou2025glean} for the three benchmark datasets. In their setting, a specified ratio of all categories is randomly selected as known categories, referred to as the known category ratio (KCR). For each known category, 10\% of the data is chosen to form the labeled dataset\footnote{\textbf{IntentGPT} \citep{rodriguez2024intentgpt} mention they use fewer than 10\% for each known category, but they do not specify the exact percentage.}, while the remaining samples constitute the unlabeled dataset. Together, these two subsets form the training dataset. In our experiments, we do not use the validation set and use only 30\% of the training dataset, restricted to the unlabeled setting with no access to labeled data. Evaluation is conducted on the same test split. 

They train an embedder using the training split and optimize it based on the labeled validation set, with the test set used solely for evaluation. In contrast, our method performs test-time embedding construction, so it requires to access test data, similar to IDAS \citet{de2023idas} and IntentGPT \citet{rodriguez2024intentgpt}, using all available text data ($D = D^{test} \cup D^u$) to construct embeddings.

Since our clustering is unsupervised, no label information is used. In practice, clustering methods require access to the text to be clustered in order to construct representations, and our approach naturally uses all available texts. 

\begin{table}[t]
\centering
\resizebox{0.48\textwidth}{!}{
\begin{tabular}{l|c|c|c|c|c}
\hline
& \# clusters & \# train & \# valid  & \# test & \# total \\
\hline
Bank77     & 77      & 9,003   & 1,000 & 3,080 & 13,083\\ 
Clinc150    &150      &  18,000  & 2,250 & 2,250 & 22,500\\ 
StackO& 20      &  18,000  & 1,000 & 1,000 & 20,500\\ 
\hline

\end{tabular}
}
\caption{Train–Validation–Test splits from previous studies \citep{an2024generalized,liang2024actively,zou2025glean}. `StackO' is the StackOverflow dataset }
\label{tab:GCDSplit}
\end{table}

\section{Selection of $d$}\label{d_select}
\begin{figure}[t]
        \centering
        \includegraphics[width=0.45\textwidth]{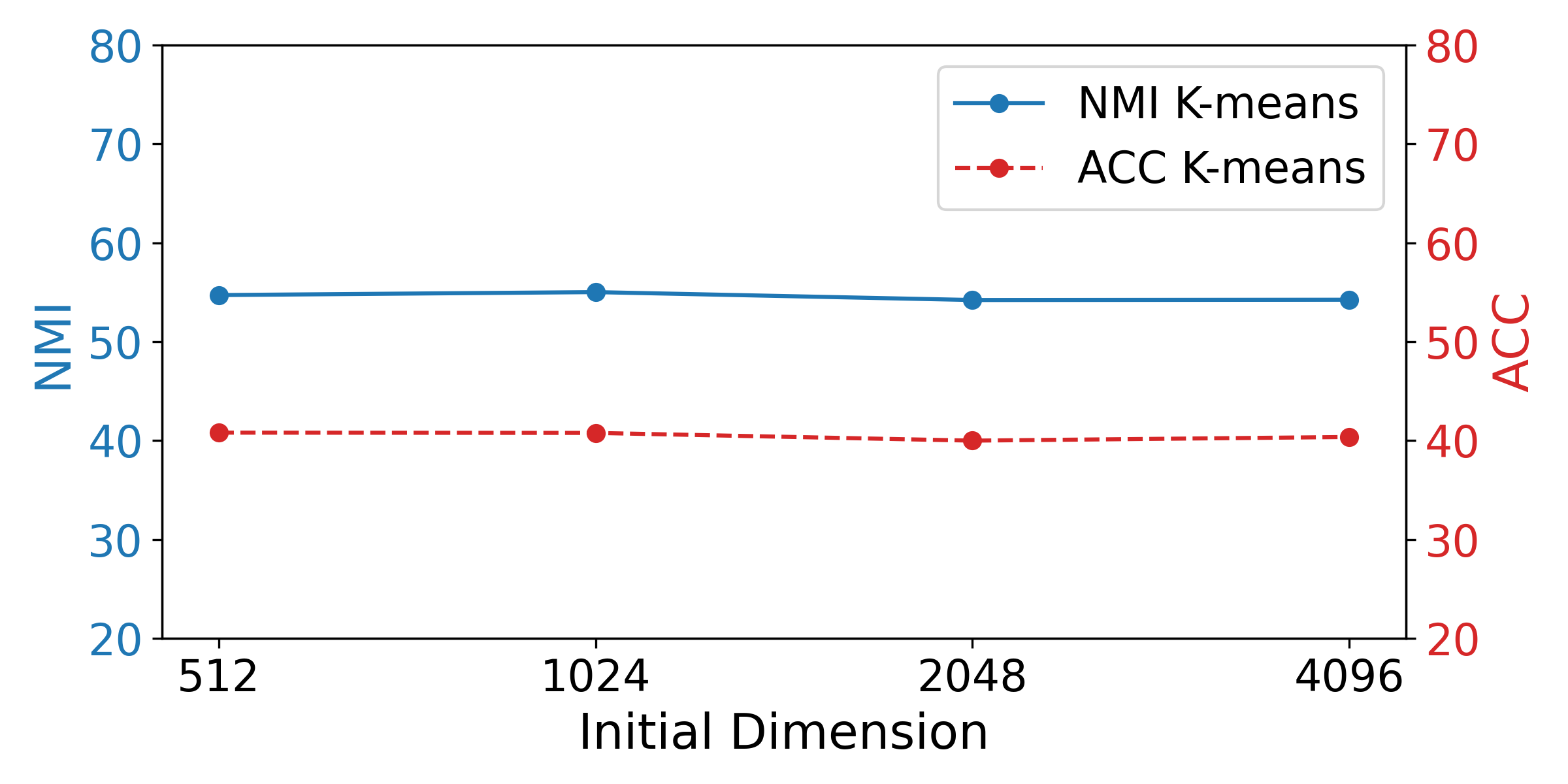}
        \caption{The average ACC and NMI across two datasets for different $d$. Results for K-means are averaged over 5 runs. $4096$ here means $\min\{4096, N\}$ because the dataset size can be smaller.}
        \label{fig:dim}
\end{figure}

gemma-2-9b-it (as LLM) and All-MiniLM-L6-v2 (as embedder) are used in the hyperparameter search. Table~\ref{tab:externaldataset} shows the two datasets that we used for selecting the dimensionality of the BoT vector $d$. Figure \ref{fig:dim} shows that they all give similar performance. We selected value $d = 1024$ and used this value across all datasets and experiments.

\begin{table}[t]
\centering
\begin{tabular}{l|c|c}
\hline
\textbf{Emerging data} & $K$ &  |$D^{test}$|\\ 

\hline
ArxivS2S     & 93     & 3,674 \\ 
Reddit       & 50     & 3,217 \\ 
\hline

\end{tabular}

\caption{External datasets used for selecting the dimensionality hyperparameter $d$}
\label{tab:externaldataset}
\end{table}








\section{Prompt} \label{llm_prompt}
Table \ref{tab:pt} shows the prompts used in our experiment. It consists of three parts: task instruction, answer format, and task. 
\begin{table}[t]
\centering
\scriptsize
\begin{tabular}{|p{7cm}|}
\hline

\textbf{CLINC150, Bank77, Mtop, and Massive} \\

\hline

\textbf{Task Instructions:} \\
Compare each Candidate Utterance to the Target Utterance. Select only Candidates with the same intent as the Target Utterance. Intent refers to the request or the purpose the user wants to achieve.\\

\hline

\textbf{GoEmo} \\
\hline

\textbf{Task Instructions:} \\

Compare each Candidate Utterance to the Target Utterance. Select only Candidates with the same emotion as the Target Utterance.
\\
\hline

\textbf{Stackoverflow} \\
\hline

\textbf{Task Instructions:} \\
Compare each Candidate Question to the Target Question. Select only Candidates with the same main programming framework, language, tool, or concept as the Target Question.
\\
\hline

\textbf{Answer Format:}\\
Only provide the final selection of Candidate Utterances by listing their numbers if they match the Target Utterance intent or request.
\begin{enumerate}[left=0pt, itemsep=0pt, topsep=0pt]
    \item If Candidates 3, 4, and 9 match, write: The Candidate utterance number(s): 3, 4, 9
    \item If no Candidates match, write: The Candidate utterance number(s): none
\end{enumerate}
Note: Stick to the answer format and avoid providing extra explanations.\\
\\
\textbf{Task:}
\\
Target Utterance: \{sentence 1\} \\
Candidate Utterances:  \\
1. \{sentence 1\}\\
... \\
m. \{sentence m\}\\
\hline

\end{tabular}
\caption{Task Instructions Prompt. Note: The boldface used here is for readability; it is not used in the prompt. If the dataset is Stackoverflow, we use `question' instead of `utterance' in the Answer format.}
\label{tab:pt}
\end{table}

\section{Effect of Different Embedders}\label{embed_analysis}

Because we use a subset instead of scanning the entire dataset for each text, the quality of the texts selected by the embedder influences the construction of our BoT vector. We report two ratios of the first iteration, averaged across all texts.

Embedder Selection Ratio:
\[
\frac{\#\{\text{\small True same-cluster in top $m$ \small as the text}\}}{m}
\]
In our experiments, we set $m = 30$. Note that Embedder here does not refer to the pre-training embeddings alone; it denotes the concatenation of the pre-training embeddings and the BoT vectors. We focus on the first iteration, i.e., $\mathbf{U}^1$, since all BoT vectors have not yet converged and they are largely evenly spaced and uninformative in the initial state. As a result, the selection relies mostly on the pre-training embeddings.

LLM Selection Ratio (LLM):
\[
\frac{
\#\{\text{\small LLM-selected items truly in the same cluster as the text}\}
}{
\#\{\text{ \small items selected by LLM}\}
}
\]
Table \ref{tab:emb_acc} shows that E5 produces the best pool on average. We also observe that the LLM ratio remains relatively stable, even when the embedder ratio fluctuates substantially.
\begin{table*}[t]

\centering
  \resizebox{0.75\textwidth}{!}{
    \begin{tabular}{lcc|lc|cc|cc|cc|cc}
    \hline
    &\multicolumn{2}{c|}{Bank 77} & \multicolumn{2}{c|}{Clinc150}  & \multicolumn{2}{c|}{Mtop} & \multicolumn{2}{c|}{Massive}& \multicolumn{2}{c|}{GoEmo} & \multicolumn{2}{c}{Average} \\
     \cline{1-3} \cline{4-5} \cline{6-7} \cline{8-9} \cline{10-11} \cline{12-13}     
    & Emb & LLM & Emb & LLM & Emb & LLM & Emb & LLM  & Emb & LLM & Emb & LLM  \\
    \hline
    \textbf{TF-IDF}& 43.26 & 68.02 & 49.71 & 82.60 & 62.93 & 78.55 & 45.28 & 70.21 & \textbf{30.85} & \textbf{36.32} & 46.41 & 67.14 \\

    \textbf{Bert}& 38.08 & 63.41 & 56.12 & 86.49 & 69.72 & 82.48 & 50.48 & 73.61 & 20.02 & 25.90 & 46.88 & 66.38 \\

    \textbf{MiniLM}& \textbf{62.03} & \textbf{75.55} & \textbf{69.76} & \textbf{89.09} & 70.88 & 81.27 & 61.58 & 76.14 & 22.33 & 27.79 & 57.32 & 69.97 \\

    \textbf{E5}& 58.64 & 73.25 & 67.80 & 88.48 & \textbf{72.00} & \textbf{81.53} & \textbf{61.66} & \textbf{76.33} & 27.97 & 32.02 & \textbf{57.61} & \textbf{70.32} \\

    \hline

  \end{tabular}
  }
  \caption{Selection Accuracy By Embedders at 1st iteration ($\%$)(\textbf{emerging data scenario)}. Bold numbers: best. LLM is Gemma.}
  \label{tab:emb_acc}

\end{table*}

\section{Percentage of Texts Converged at Iteration $t$}\label{convergence}
\begin{figure}[t]
        \centering
        \includegraphics[width=0.45\textwidth]{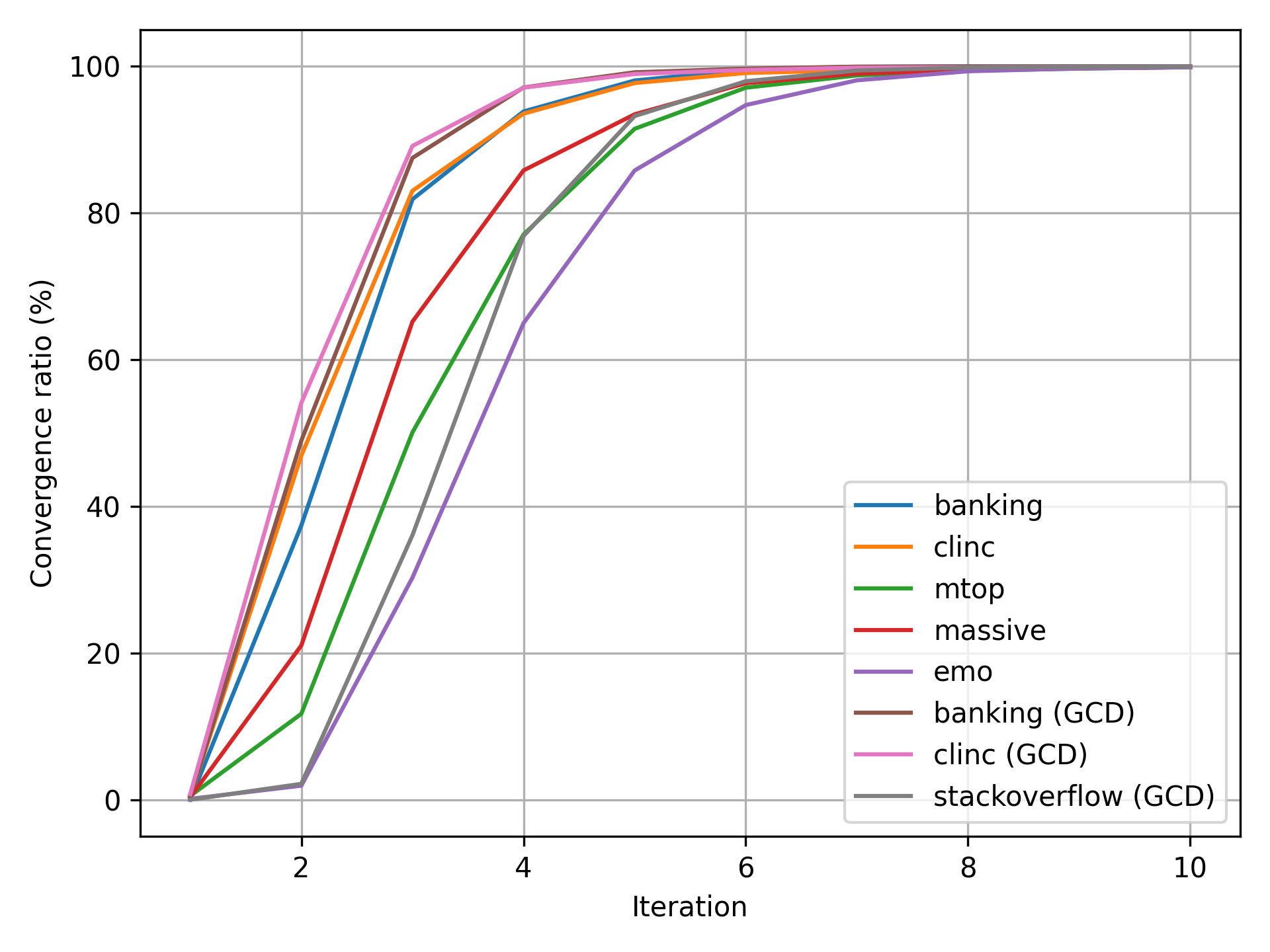}
          \caption{Ratio of texts converged at iteration $t$. All-MiniLM-L6-v2 (as embedder) and Qwen3-8B (as LLM) were used.}
        \label{fig:iter_con}
\end{figure}

Figure \ref{fig:iter_con} shows ratio of text convergence. It shows most datasets have 90\% convergence ratio by the 5th iteration, and by the 7th iteration, all datasets achieve a  95\% convergence ratio.

\begin{table*}[t]
  \centering
  \setlength{\tabcolsep}{2.5pt}
  \footnotesize
  \resizebox{\textwidth}{!}{
    \begin{tabular}{lcc|lc|cc|cc|cc|cc|cc|cc|cc}
    \hline
    &\multicolumn{2}{c|}{Bank 77} & \multicolumn{2}{c|}{Clinc150}  & \multicolumn{2}{c|}{Mtop} & \multicolumn{2}{c|}{Massive}& \multicolumn{2}{c|}{GoEmo} & \multicolumn{2}{c|}{Bank 77\textsubscript{GCD}} & \multicolumn{2}{c|}{Clinc150\textsubscript{GCD}}  & \multicolumn{2}{c|}{StackO\textsubscript{GCD}}& \multicolumn{2}{c}{Average} \\
     \cline{1-3} \cline{4-5} \cline{6-7} \cline{8-9} \cline{10-11} \cline{12-13} \cline{14-15} \cline{16-17}\cline{18-19}
    & NMI & Acc & NMI & Acc & NMI & Acc& NMI & Acc & NMI & Acc & NMI & Acc & NMI & Acc & NMI & Acc& NMI & Acc\\
    \hline

    Gemma & 85.44 & 73.12 & 94.45 & 87.58 & 72.18 & 39.40 & 76.74 & 66.84 & 21.77 & 25.22 & 86.90 & 74.93 & 95.32 & 88.71 & 81.21 & 82.54 & 76.75 & 67.29 \\
    
    Qwen & 85.01 & 71.63 & 93.79 & 87.12 & 73.29 & 37.45 & 74.24 & 61.47 & 21.61 & 25.83 & 86.53 & 74.31 & 95.61 & 89.41 & 82.15 & 83.42 & 76.53 & 66.33 \\ 
    
    Llama & 81.75 & 66.18 & 91.66 & 83.12 & 70.12 & 35.79 & 73.26 & 60.40 & 18.91 & 20.81 & 83.16 & 68.16 & 94.00 & 85.66 & 77.42 & 80.10 & 73.78 & 62.53 \\

    \hline

  \end{tabular}
  }

  \caption{Further analysis of results using different LLMs. }

  \label{tab:all_LLM_results}
\end{table*}

\section{Results by Different LLMs}\label{LLM_result_by_dataset}
We report the detailed results of all datasets by different LLMs in Table \ref{tab:all_LLM_results}. Qwen and Gemma consistently outperform LLaMA. We also report the LLM selection accuracy (see Appendix \ref{embed_analysis} for the definition). Table \ref{tab:all_LLM_analysis} shows that Gemma and Qwen have similar capabilities in identifying similar texts, outperforming LLaMA. This finding aligns with prior work \cite{lin2025spill}.

\begin{table*}[t]
  \centering
  \setlength{\tabcolsep}{2.5pt}
  \footnotesize
  {\small
    \begin{tabular}{lcc|lc|cc|cc|cc|cc}
    \hline
    &\multicolumn{2}{c|}{Bank 77} & \multicolumn{2}{c|}{Clinc150}  & \multicolumn{2}{c|}{Mtop} & \multicolumn{2}{c|}{Massive}& \multicolumn{2}{c|}{GoEmo} &  \multicolumn{2}{c}{Average} \\
     \cline{1-3} \cline{4-5} \cline{6-7} \cline{8-9} \cline{10-11} \cline{12-13} 
    & Emb & LLM & Emb & LLM & Emb & LLM & Emb & LLM  & Emb & LLM & Emb & LLM  \\
    \hline

    Gemma & \textbf{62.03} & \textbf{75.55} & 69.76 & 89.09 & 70.88 & 81.27 & \textbf{61.58} & 76.14 & \textbf{22.33} & 27.79 & 57.32 & 69.97 \\
    
    Qwen & 61.08 & 75.32 & \textbf{70.16} & \textbf{90.14} & \textbf{72.44} & \textbf{84.19} & 60.88 & \textbf{76.62} & 22.27 & \textbf{28.91} & \textbf{57.36} & \textbf{71.04} \\
    
    Llama & 56.15 & 64.57 & 64.05 & 78.95 & 67.99 & 75.16 & 58.17 & 67.74 & 19.96 & 23.43 & 53.26 & 61.97 \\

    \hline

  \end{tabular}
  }

  \caption{Selection Accuracy By LLMs at 1st iteration($\%$)(\textbf{emerging data scenario)}. Bold numbers: best. All-MiniLM-L6-v2 used as backbone embedder. }

  \label{tab:all_LLM_analysis}
\end{table*}

\section{HDBSCAN Clustering}\label{HDBSCAN}

To reflect real-world settings with unknown $K$, we apply HDBSCAN \citep{mcinnes2017hdbscan} to partition $D^{test}$ into $\hat{K}$ clusters. We set \texttt{min\_samples = 1} to reduce noise. The noise points are assigned to the nearest cluster centroid. We perform an incremental grid search over \texttt{min\_cluster\_size} values of $10, 15, \ldots, 50$, stopping early if the silhouette score does not improve \citep{rousseeuw1987silhouettes}. Using the silhouette score for hyperparameter search follows prior work \citep{gung2023intent}.

Tables \ref{tab:EmgH} and \ref{tab:GCDH} shows the detailed evaluation result with HDBSCAN clustering.

\begin{table*}[t]

\centering
  \resizebox{0.75\textwidth}{!}{
    \begin{tabular}{lcc|lc|cc|cc|cc|cc}
    \hline
    &\multicolumn{2}{c|}{Bank 77} & \multicolumn{2}{c|}{Clinc150}  & \multicolumn{2}{c|}{Mtop} & \multicolumn{2}{c|}{Massive}& \multicolumn{2}{c|}{GoEmo} & \multicolumn{2}{c}{Average} \\
     \cline{1-3} \cline{4-5} \cline{6-7} \cline{8-9} \cline{10-11} \cline{12-13}     
    & NMI & Acc & NMI & Acc & NMI & Acc& NMI & Acc & NMI & Acc & NMI & Acc \\
    \hline
    \textbf{TF-IDF}&&&&&&&&&&&\\
    Plain & 53.06 & 33.34 & 56.55 & 32.64 & 39.79 & 22.83 & 41.26 & 26.09 & 11.56 & 16.10 & 40.44 & 26.20 \\

    LeBoT (Ours) & \textbf{79.92} & \textbf{65.23} & \textbf{89.29} & \textbf{79.58} & \textbf{71.14} & \textbf{46.62} & \textbf{69.81} & \textbf{56.46} & \textbf{30.93} & \textbf{24.47} & \textbf{68.22} & \textbf{54.47} \\
    \hline
    \textbf{Bert}&&&&&&&&&&&\\

    Plain & 0.84 & 1.69 & 73.03 & 50.87 & 63.95 & 40.97 & 2.86 & 8.18 & 1.35 & 12.05 & 28.41 & 22.75 \\

    LeBoT (Ours)  & \textbf{76.12} & \textbf{57.40} & \textbf{93.07} & \textbf{85.62} & \textbf{75.94} & \textbf{51.64} & \textbf{73.71} & \textbf{63.97} & \textbf{20.78} & \textbf{15.68} & \textbf{67.92} & \textbf{54.86} \\

    \hline

    \textbf{MiniLM}&&&&&&&&&&&\\

    Plain & 78.78 & 60.39 & 86.79 & 67.71 & 67.30 & 55.91 & 62.55 & 48.15 & 11.31 & 18.62 & 61.35 & 50.16 \\

    LeBoT (Ours)   & \textbf{85.13} & \textbf{71.85} & \textbf{94.52} & \textbf{87.58} & \textbf{76.83} & \textbf{59.51} & \textbf{76.88} & \textbf{67.78} & \textbf{23.18} & \textbf{21.92} & \textbf{71.31} & \textbf{61.73} \\

    \hline
    \textbf{E5}&&&&&&&&&&&\\
    
    Plain & 78.68 & 62.18 & 87.57 & 65.62 & 63.45 & \textbf{53.79} & 1.15 & 7.74 & 15.43 & 17.70 & 49.26 & 41.41 \\

    LeBoT (Ours)   & \textbf{85.03} & \textbf{70.65} & \textbf{93.82} & \textbf{85.80} & \textbf{75.38} & 53.17 & \textbf{76.33} & \textbf{62.59} & \textbf{28.07} & \textbf{29.59} & \textbf{71.73} & \textbf{60.36} \\
    
    \hline

  \end{tabular}
  }
  \caption{Five-benchmark results (\textbf{emerging data scenario, HDBSCAN}). Gemma is used in LeBoT. \textbf{Plain}: direct embedding clustering. Bold numbers: best within the embedder. }
  \label{tab:EmgH}

\end{table*}

\begin{table}[t]
  \centering
  \setlength{\tabcolsep}{2.5pt}
    \resizebox{0.49\textwidth}{!}{
    \begin{tabular}{l|cc|cc|cc|cc}
    \hline
     & \multicolumn{2}{c|}{Bank 77} & \multicolumn{2}{c|}{Clinc150}  & \multicolumn{2}{c|}{StackO}& \multicolumn{2}{c}{Average} \\
     \cline{2-3} \cline{4-5} \cline{6-7} \cline{8-9}    
    Method & NMI & Acc & NMI & Acc & NMI & Acc & NMI & Acc \\
    \hline
    Plain& 78.78 & 60.39 & 83.02 & 51.82 & 66.26 & 57.20 & 76.02 & 56.47 \\

    LeBoT & \textbf{86.75} & \textbf{74.64} & \textbf{95.24} & \textbf{88.36} & \textbf{81.43} & \textbf{83.50} & \textbf{87.81} & \textbf{82.17} \\
    \hline

  \end{tabular}
  }

    \caption{Three-benchmark results (\textbf{GCD scenario, HDBSCAN}) Gemma is used in LeBoT. Bold numbers: best. All-MiniLM-L6-v2 used as backbone embedder. \textbf{Plain}: direct embedding clustering.}

  \label{tab:GCDH}
\end{table}

Table \ref{tab:EM_Khat} and Table \ref{tab:GCD_Khat} show the estimated number of clusters, $\hat{K}$, for the two scenarios. The results indicate that the refined embeddings produced by our method are generally closer to the ground truth than the pre-trained (Plain) embeddings. Except for the Bank77 and GoEmo datasets.
\begin{table}[t]
  \centering
  \setlength{\tabcolsep}{2.5pt}
  \small{
    \begin{tabular}{l|c|c|c|c|c}
    \hline
    & Bank 77 & Clinc150 & Mtop & Massive & GoEmo  \\
    \hline
    \textbf{Bert} & & & & \\
    Plain & 2 & 111 & 62 & 2 & 2 \\
    LeBoT & 69 & 144 & \textbf{77} & \textbf{61} & 51 \\
    \hline
    \textbf{MiniLM} & & & & \\
    Plain & \textbf{78} & 139 & 17 & 27 & 11 \\
    LeBoT & 66 & 145 & 67 & 56 & 53 \\
    \hline
    \textbf{E5} & & & & \\
    Plain & 73 & 134 & 9 & 2 & \textbf{3} \\
    LeBoT & 68 & 147 & 64 & 63 & 60 \\
    \hline

  \end{tabular}
  }

  \caption{\textbf{Emerging} scenario: Estimated Number of Clusters $\hat{K}$ by HDBSCAN. \textbf{gemma-2-9b-it} used in LeBoT.}

  \label{tab:EM_Khat}
\end{table}

\begin{table}[t]
  \centering
  \setlength{\tabcolsep}{2.5pt}
  \small{
    \begin{tabular}{l|c|c|c}
    \hline
    & Bank 77 & Clinc150 & Stackoverflow  \\
    \hline
    Plain & \textbf{78} & 82 & 13 \\
    LeBoT & 80 & \textbf{139} & \textbf{19} \\
    \hline
    True $K$ & 77& 150& 20\\    
    \hline

  \end{tabular}
  }

  \caption{\textbf{GCD} scenario: Estimated Number of Clusters $\hat{K}$ by HDBSCAN. All-MiniLM-L6-v2 used as backbone embedder. \textbf{gemma-2-9b-it} was used in LeBoT.}

  \label{tab:GCD_Khat}
\end{table}

\section{Details of the full dataset clustering}\label{simu_dim}

\paragraph{Datasets Used with LLM Guidance}

Table \ref{tab:fulldataset} shows the number of examples we use with the LLM. We re-use the representations derived in the emerging data setting, so the number of examples with LLM guidance remains the same as in Table \ref{tab:dataset}, i.e., $|D^{test}|$, and there is no need to re-run all experiments. The examples account for 20.6\% to 23.5\% of datasets. Examples with LLM guidance are used to train the MLP, since their BoT vectors have been derived. The pre-trained embeddings from all-MiniLM-L6-v2 serve as input, and the LLM-generated BoT vectors serve as target output for supervised learning with an MLP. 20\% of the examples are used as a validation set to tune the hyperparameters.

\paragraph{MLP Architecture and Hyperparameter Search}

We build an MLP for each dataset. We only tune the hidden layer dimensionality using the validation set, with search over $\{512, 1024, 2048\}$. 

\paragraph{Clustering of the entire dataset with MLP output}
We first encode all examples into pre-trained embeddings using all-MiniLM-L6-v2, then use the trained MLP to derive BoT versions of these embeddings. These BoT vectors are used to cluster all examples.

\begin{table}[H]
\centering
\resizebox{0.5\textwidth}{!}{
\begin{tabular}{l|c|c|c|c|c}
\hline
Dataset & $K$ & Total   & LLM & No-LLM &  LLM ratio \\
\hline
Bank77     & 77      & 13,083  & 3,080 & 10,003  & 23.5\%\\ 
Clinc150    &150      &  22,500  & 4,500 & 18,000& 20\%\\ 
\hline

\end{tabular}
}
\caption{Dataset Statistics. }
\label{tab:fulldataset}
\end{table}

\paragraph{Evaluation}
For HDBSCAN, we set \texttt{min\_samples = 1} to reduce noise. The noise points are assigned to the nearest cluster centroid. We perform an incremental grid search over \texttt{min\_cluster\_size} values of $50, 75, 100, 125, \ldots, 250$ -- five times the original range -- stopping early if the silhouette score does not improve. For K-means, we set the number of clusters equal to the ground-truth cluster count $K$.

\end{document}